\documentclass[lettersize,journal]{IEEEtran}
\newtheorem{remark}{Remark}
\usepackage{hyperref}
\hypersetup{hidelinks, colorlinks=true, allcolors=blue, pdfstartview=Fit, breaklinks=true}
\usepackage{flushend}
\usepackage{amsmath} 
\usepackage{graphicx}
\usepackage{subcaption}
\usepackage{xcolor}
\usepackage{tikz}
\usepackage{cite}
\usepackage{algorithm} 
\usepackage[noend]{algorithmic}
\usepackage[numbers,sort&compress]{natbib}
\usepackage{amsmath}
\usepackage{amssymb}
\usepackage{multirow}
\usepackage{multicol}
\usepackage{xcolor}
\usepackage{listings}
\floatname{algorithm}{Procedure}
\usepackage{flushend}
\usepackage{amsmath} 
\usepackage{graphicx}
\usepackage{subcaption}
\usepackage{xcolor}
\usepackage{tikz}
\usepackage{cite}
\usepackage{algorithm} 
\usepackage[noend]{algorithmic}
\usepackage{amsmath}
\usepackage{amssymb}
\usepackage{multirow}
\usepackage{multicol}
\usepackage{xcolor}
\usepackage{listings}
\usepackage{enumitem}
\usepackage{stfloats}
\floatname{algorithm}{Procedure}

\title{\LARGE \bf
BT-TL-DMPs: A Novel Robot TAMP Framework  Combining Behavior Tree, 
Temporal Logic and Dynamical Movement Primitives
}

\author{Zezhi Liu$^{1}$, Shizhen Wu$^{1}$,  Hanqian Luo$^{2}$, Deyun Qin$^{1}$, and Yongchun Fang$^{1}$
\thanks{This work was supported by National Natural Science Foundation of China under Grant 62233011. (Zezhi Liu and Shizhen Wu are co-first authors)
(Corresponding Author: Yongchun Fang) }
\thanks{$^{1}$Zezhi Liu, Shizhen Wu, Deyun Qin and Yongchun Fang are with the Institute of Robotics and Automatic Information Systems, College of Artificial Intelligence, Nankai University, Tianjin 300350, China, and also with the Tianjin Key Laboratory of Intelligent Robotics, Nankai University, Tianjin 300350, China (e-mail: zezhi.liu@mail.nankai.edu.cn, szwu@mail.nankai.edu.cn, qindy@mail.nankai.edu.cn, fangyc@nankai.edu.cn).}%
\thanks{$^{2}$Hanqian Luo is with the College of Artificial Intelligence, Nankai University, Tianjin 300350, China, and also with the Department of Computing, The Hong Kong Polytechnic University, Hong Kong, China (e-mail: hanqian.luo@connect.polyu.hk).}%
}

\lstdefinestyle{pddl}{
    basicstyle=\ttfamily\small,
    columns=fullflexible,
    keepspaces=true,
    showstringspaces=false,
    keywordstyle=\color{blue},
    morekeywords={domain, action, parameters, precondition, effect, task_requirements, motion_constraints},
    literate={-}{-}1, 
}

\begin{document}
\maketitle

\begin{abstract}

In the field of Learning from Demonstration (LfD), enabling robots to generalize learned manipulation skills to novel scenarios for long-horizon tasks remains challenging. Specifically, it is still difficult for robots to adapt the learned skills to new environments with different task and motion requirements, especially in long-horizon, multi-stage scenarios with intricate constraints. This paper proposes a novel hierarchical framework, called BT-TL-DMPs, that integrates Behavior Tree (BT), Temporal Logic (TL), and Dynamical Movement Primitives (DMPs) to address this problem. Within this framework, Signal Temporal Logic (STL) is employed to formally specify complex, long-horizon task requirements and constraints. These STL specifications are systematically transformed to generate reactive and modular BTs for high-level decision-making task structure. An STL-constrained DMP optimization method is proposed to optimize the DMP forcing term, allowing the learned motion primitives to adapt flexibly while satisfying intricate spatiotemporal requirements and, crucially, preserving the essential dynamics learned from demonstrations. The framework is validated through simulations demonstrating generalization capabilities under various STL constraints and real-world experiments on several long-horizon robotic manipulation tasks. The results demonstrate that the proposed framework effectively bridges the symbolic-motion gap, enabling more reliable and generalizable autonomous manipulation for complex robotic tasks. The project website can be found at: \href{https://nkrobotlab.github.io/BT-TL-DMPs/}{https://nkrobotlab.github.io/BT-TL-DMPs/}

\end{abstract}

\section{INTRODUCTION}

Robots are increasingly expected to perform complex, long-horizon tasks that require reasoning across multiple levels of abstraction. Human individuals naturally leverage multi-modal information—from high-level task context to low-level motor execution—to generalize manipulation skills across diverse scenarios. Emulating such capabilities for robots remains a crucial goal in the field of autonomous manipulation.

One promising direction is \emph{Task and Motion Planning} (TAMP)~\citep{garrett2021integrated, guo2023recent, zhao2024survey}, which integrates symbolic task planning with continuous motion planning. TAMP enables robots to generate feasible action sequences that bridge abstract goals and executable motor commands, supporting long-horizon task execution in structured environments. In parallel, methods from \emph{Learning from Demonstration} (LfD)~\citep{ravichandar2020recent, lu2021constrained}, particularly those based on Dynamical Systems (DS), offer a way for robots to acquire human-like manipulation behaviors by observing expert demonstrations, reducing the need for complicated explicit programming.

However, several challenges remain when bridging TAMP and LfD into a unified framework. First, reactive LfD methods augmented with finite-state machines (FSM), e.g., combining DMPs with HMMs in \citep{zhou2020incremental}, ProMPs with FSMs and  Control Barrier Functions (CBFs) in \citep{davoodi2022rule}, or DMPs with Buchi Automata in \citep{nawaz2024reactive}, can handle discrete task switching and safety yet often lack the flexibility to adapt continuous motion primitives to new scenarios. Second, although behavior trees offer enhanced modularity and reactivity when compared with FSMs \citep{colledanchise2018behavior, ogren2022behavior,iovino2024comparison}, existing work yet need to demonstrate how to simultaneously learn or generate both the discrete subtask hierarchy and the underlying continuous controllers \cite{cloete2024adaptive}. Third, while Signal Temporal Logic (STL) has been successfully used as a specification interface in optimization-based TAMP \citep{zhao2024survey,leung2023backpropagation} and Large Language Model (LLM) based planning \citep{chen2023nl2tl,wang2024chatstl}, its integration with DS-based motion planners remains sparse: the only reported approach encodes STL as time-varying CBFs to filter outputs of DS-based planner \citep{nawaz2024reactive}. At the same time, CBF methods often fail on complex specifications or yield infeasible solutions \citep{yin2024formal}. These gaps motivate us to develop a new task–motion interface that can (i) propagate rich symbolic and temporal constraints directly into learned motion primitives, and (ii) leverage STL in a way that preserves feasibility and robustness across diverse, long-horizon tasks. More reviews can be found in Sec. \ref{related}.

\begin{figure*}[!htb]
        \centering
        \includegraphics[width=0.85\hsize]{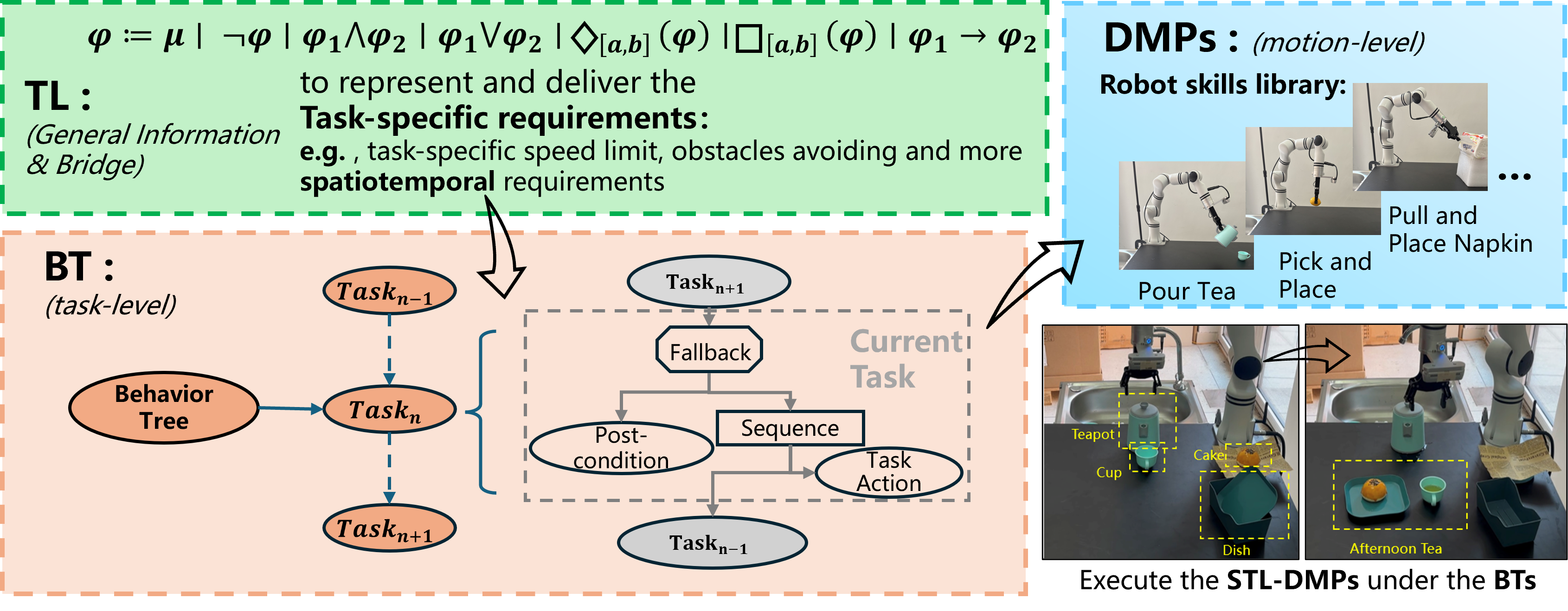}
        \caption{The proposed robot TAMP framework which is combined with BT, TL and DMPs.}
        \label{fig-1}
\end{figure*}

To address these challenges, a hierarchical planning framework is proposed that integrates task‐level (BTs) and motion‐level (DMPs) planning to reuse and generalize learned skills across diverse scenarios effectively, as shown in Fig.~\ref{fig-1}.
This framework ensures that the robot can not only plan the tasks but also translate them into executable actions while maintaining necessary constraints across different environments. Through this approach, robots are equipped with a robust mechanism to make task-level decisions and generalize learned behaviors efficiently, fostering the ability to tackle a wide range of complex tasks in real-world settings.
More specifically, the main contributions of this paper can be summarized as follows:
\begin{itemize}

   \item [1)]   {A task-level planning method (from STL to BTs) is designed, enabling the soundness of high-level decisions.}
   

    \item [2)] An STL‐constrained DMP optimization method is developed to preserve demonstrated dynamics and enforce task‐specific spatiotemporal constraints for efficient skill adaptation and generalization.

    \item [3)] Simulation and experiments are  conducted to evaluate the proposed framework, providing empirical validation of its effectiveness.
\end{itemize}


\subsection{Related works}\label{related} 
\subsubsection{Ways to Increase the Reactive Ability of LfD} 

Recently, researchers have attempted to improve the reactive ability of LfD by finite-state machine, such as \citep{zhou2020incremental, davoodi2022rule, nawaz2024reactive}.   
In detail, in  \citep[Chapter 3]{zhou2020incremental}, DMPs are combined with a finite-state machine (FSM) for  task representation, and a Bayesian nonparametric Hidden Markov Model (HMM) is applied for robot movement recognition. In \citep{davoodi2022rule}, 
  an FSM is incorporated with ProMPs to represent the whole complex task in the environment, and  
 control barrier function-based methods are utilized to execute the learned trajectories safely.
More recently,  a modular reactive planner that integrates discrete linear temporal logic-based Buchi Automaton (i.e., one kind of FSM) and continuous LfD-based motion plans is proposed, so as to guarantee task satisfaction and stability while adapting to unforeseen disturbances and external observations \citep{nawaz2024reactive}.

{The Behavior Tree (BT) methods have been proposed to substitute FSMs in decision-making to enhance the reactivity and modularity \citep{colledanchise2018behavior,ogren2022behavior,iovino2023programming,iovino2024comparison}. The main advantage of behavior trees
is to enable the low-level controller to arrange the hierarchy of sub-tasks in a discrete space and plan continuous actions among these discrete sub-goals more optimally. BT methods can not only achieve the best performance on overall task execution under different contexts, but also obtain more generalized and adaptive behaviors for solving a class of tasks that share common features, i.e., bin-picking, handling, and sorting. Due to such advantages, recently, researchers have noticed the idea of augmenting DMPs with BTs \cite{wang2024novel, dominguez2025beyond}. 
In \cite{wang2024novel}, a BT is manually constructed from task-level switching logic and steps to guide transitions between different DMPs. In \cite{dominguez2025beyond}, by integrating DMP controllers into a BT framework, the BT structure and DMP actions are learned jointly from single demonstrations. 
 In comparison, our approach focuses on generating BTs from specific Signal Temporal Logic formulas and modifying the underlying DMPs accordingly to satisfy some STL formula.}

\subsubsection{The Roles of  STL in TAMP} 

 Temporal Logic (TL) specifications have become a popular tool for formulating what robots need to do in a task without requiring information on how to complete the task in the control and robotics community \citep{plaku2015motion}. 
 In contrast to classical specification languages, such as PDDL \citep{aeronautiques1998pddl}, TLs are rich and concise, often closer to resemble natural language and can be used to specify various temporal events of tasks without explicitly modeling the robot’s action space.  
STL is defined over continuous signals and real predicates, providing quantitative semantics, 
 that indicates the margin of satisfaction or violation of the given specification.  Due to such advantages, 
STL plays an important part in the area of TAMP, as reviewed in recent survey papers \citep{garrett2021integrated,
guo2023recent,
zhao2024survey}, where 
STL has been widely applied in optimization-based trajectory planning \citep{zhao2024survey, leung2023backpropagation} and more recently, STL has emerged as a key interface for connecting high-level natural language instructions with low-level motion planning in LLM-based robotic planning frameworks \citep{chen2023nl2tl, wang2024chatstl, van2024vernacopter}.

There are still a few works to consider the idea of integrating STL with DMP. 
As far as we know,  the recent reference \citep{nawaz2024reactive} is the only work that discusses how to augment DMP (i.e., a DS-based motion planner)  with STL.
In detail, time-critical tasks such 
as finite-time reach tasks are modeled as STL specifications, then following in line with the procedure given in \citep{lindemann2018control}, the STL specification is encoded as a time-varying CBF, based on which a quadratic program-based safety filter is established to minimally modify the output of a DS-based motion planner. However, it is remarked in the latest survey paper \citep[Sec. 4.3]{yin2024formal} that CBF-based approaches are generally tailored to a simpler class of STL tasks, and for more complex STL tasks, this approach may potentially yield no solution even when a feasible one exists. Such observation motivates us to find an alternative method to generalize DMP with STL.

\section{PRELIMINARIES} \label{PRE}
This work leverages Dynamic Movement Primitives for learning from demonstration, Temporal Logics for task specification and evaluation, and Behavior Trees for modular task execution. More specifically,  DMPs model complex pose trajectories from demonstrations, 
enabling flexible and generalizable motion reproduction. STL provides a formal framework for expressing and quantitatively evaluating temporal specifications over continuous signals. BTs offer a reactive and hierarchical structure for behavior composition and decision-making. Detailed formulations and mathematical backgrounds for DMPs, STL, and BTs are provided in the following part.

\subsection{Learning from Demonstration}\label{subsec:LfD}
\textbf{Dynamic Movement Primitives (DMPs)} are a set of second-order nonlinear autonomous equations inspired by biological motor primitives \citep{saveriano2023dynamic}, 
which is a versatile tool in robotics, allowing for flexible adaptation to various tasks and environmental conditions, as well as providing the foundation for learning-based approaches in movement generation.
The basic theory of DMPs is reviewed here. 
To represent the position of robot’s end effector $\mathbf{y}$, the traditional DMPs are given by:
\begin{equation}\label{eq:DMPs}
\begin{aligned}
 \tau \dot{\mathbf{v}} &= \alpha (\beta ( \mathbf{y}_g - \mathbf{y}) - {\mathbf{v}}) + \mathbf{F}(\phi), \\
     \tau \dot{\mathbf{y}} &= \mathbf{v},
\end{aligned}
\end{equation}
where \( \tau \in \mathbb{R}\) is a time constant, 
the signals \( \dot{\mathbf{y}}  \in \mathbb{R}^{3} \) and \( \ddot{\mathbf{y}} \in \mathbb{R}^{3}  \) represent velocity and acceleration, while  \( \mathbf{y} \)  and  \( \mathbf{y}_g \) denotes the actual and goal position respectively.
\( \alpha \) and \( \beta \) are the elastic coefficient and damping coefficient of the spring-damper system, respectively. The function \( \mathbf{F}(\phi) \) introduces a phase-dependent forcing term that allows the system to learn and repeat its behavior over time. The phase variable \( \phi \) decreases from $1$ to $0$ over time, causing \(  \mathbf{F}(\phi) \) to decay and ensuring convergence towards the goal.

{In this paper, multiple demonstrations are used when learning from demonstrations, the GMM-GMR method is utilized to obtain the mean \(\hat\mu_p(\phi)\) and variance \(\hat\Sigma_p(\phi)\) of the demonstration trajectories,} which are then used to weight and regularize the DMP forcing‐term optimization in Section \ref{subsec:3.3}. 
The mean trajectory $\hat{\mu}_p(\phi)$, representing the learned trajectory of DMP, is rewritten as $\mathbf{y}_\text{demo}(\phi)$ for clarity, and used as shown in \eqref{eq:DMP_learn}.

The learned forcing term $\mathbf{F}$ of DMPs in the system \eqref{eq:DMPs} is designed as $\mathbf{F}_\text{lrn}$:
\begin{align}
\mathbf{F}_\text{lrn}(\phi) &= \frac{\sum w_i\Phi_i(\phi)}{\sum \Phi_i(\phi)} r ,
        \label{eq0.4} \\
 \Psi_i(\phi) &= \exp(h_i(\cos(\phi-c_i)-1)),        \label{eq0.5}
\end{align}
where the equations show that the forcing term are composed by basis functions $\Phi$ and the corresponding weights $w$. The parameters of the basis functions are setted in advance, only the weight parameters $\omega_i$ are learned from the demonstrations:
\begin{equation}
    \mathbf{F}_\text{lrn} = \tau^2 \ddot{\mathbf{y}}_\text{demo} - \alpha (\beta (\mathbf{y}_g - \mathbf{y}_\text{demo}) - \tau \dot{\mathbf{y}}_\text{demo}).
\label{eq:DMP_learn}
\end{equation}
Once the parameters are learned, DMPs can reproduce the demonstrated movements and generalize them to new environments by modifying the coupling terms and incorporating constraints \cite{saveriano2023dynamic, liu2025novel}.


To represent the orientation of the robot end effector, the quaternion-based DMPs can be expressed as:
\begin{equation}
        \label{eq:QDMPs}
        \begin{aligned}
        &\tau_{q} \dot{\boldsymbol{\eta}} = \alpha_q(\beta_q 2 \log (\mathbf{q}_{g} * \mathbf{\bar{q}}) - \boldsymbol{\eta}) + \mathbf{F}_q(\phi_q), 
        \\
        & \tau_{q}  \dot{\mathbf{q}} = \frac{1}{2} \boldsymbol{\eta} * \mathbf{q},
        \end{aligned}
\end{equation}
where $\tau_{q} $ remains the time constant. The quaternion $\mathbf{q} \in \bf{S}^3\subset \mathbb{R}^{4}$, a commonly used rotation representation, is written as $\mathbf{q} = v + \mathbf{u}  $, where $v \in \mathbb{R}$ and $\mathbf{u} \in \mathbb{R}^3$.  The conjugate of the quaternion is denoted as $\mathbf{\bar{q}} = v - \mathbf{u}$. 
Similarly, $\alpha_q$ and $\beta_q$ are the spring-like damping and elastic coefficients, respectively. 
The term $\mathbf{F}_q$ represents the forcing term within quaternion space. The term $2\log (\mathbf{q}_{g} * \mathbf{\bar{q}})$ in equation \eqref{eq:QDMPs},  analogous to $(\mathbf{y}_{g} - \mathbf{y})$ in traditional DMPs \eqref{eq:DMPs}, represents the angular velocity required to rotate from quaternion $\mathbf{q}$ to $\mathbf{q}_g$, pulling the trajectory towards the goal point \citep{saveriano2023dynamic}.

\subsection{Behavior Tree}\label{sec:BT}

\textbf{Behavior Tree (BT)} is a modular, reactive method for task planning in robotics and AI, offering greater flexibility and reuse than Finite State Machines (FSMs) \citep{colledanchise2018behavior,ogren2022behavior,iovino2024comparison}. As in \citep{colledanchise2018behavior}, each BT can be expressed as a controller–status pair:
\begin{align}
    \mathcal{T}_i = \bigl(u_i(x),\,r_i\bigr),
\end{align}
where \(u_i(x):\mathcal{X}\to\mathcal{U}\) maps a robot state \(x\in\mathcal{X}\subset\mathbb{R}^n\) to control input \(u\in\mathcal{U}\subset\mathbb{R}^m\), and \(r_i:\mathcal{X}\to\{\mathcal{R},\mathcal{S},\mathcal{F}\}\) returns Running (\(\mathcal{R}\)), Success (\(\mathcal{S}\)), or Failure (\(\mathcal{F}\)). 
Nodes in a BT are of two kinds: \emph{Control flow}: Sequence, Fallback, Parallel, Decorator; and \emph{Execution}: Action, Condition.
\noindent\textbf{Sequence:} Ticks children \(C_1,\dots,C_l\) in order, returning the first \(\mathcal{F}\) or \(\mathcal{R}\), returns \(\mathcal{S}\) if all succeed.  
\textbf{Fallback:} Ticks left to right, returning on the first \(\mathcal{S}\) or \(\mathcal{R}\), returns \(\mathcal{F}\) only if all fail.  
\textbf{Parallel:} Aggregates children’s statuses: succeeds when \(\ge\alpha\) report \(\mathcal{S}\), fails when \(>\beta\) report \(\mathcal{F}\), else runs. More details can be found in \citep{colledanchise2018behavior}.

\subsection{Signal Temporal Logic}\label{sec:STL}

\textbf{Signal Temporal Logic (STL)} is a formalism used to specify temporal properties of continuous signals over time, making it particularly suitable for reasoning about the behavior of dynamical systems in control, robotics, and verification tasks \citep{guo2023recent,zhao2024survey, meli2023logic}. STL combines standard temporal logic with predicates over continuous signals, allowing the specification of requirements such as ``the robot should reach a certain position within a specified time window" or ``a constrain should remain within a given range for a certain duration".

Mathematically, STL formulas are constructed from a set of atomic propositions, logic operators, and temporal operators. A basic STL formula $\varphi$ can be recursively defined as follows \citep{leung2023backpropagation, yu2024continuous}: 
\begin{equation}\label{eq::stl_formula}
\varphi \! := \! \mu  \!  \mid   \! \neg \varphi  \!  \mid  \!  \varphi_1 \wedge \varphi_2  \!  \mid  \!  \varphi_1 \vee \varphi_2  \!  \mid   \! \Diamond_{[a,b]}(\varphi)  \!  \mid  \!  \square_{[a,b]}(\varphi)   \!   \mid  \!  \varphi_1 \rightarrow \varphi_2, 
\end{equation}
where  $\mu$ is  an atomic proposition, defined by a predicate over continuous signals $\xi(t): \mathbb{R} \to \mathbb{R}^n$.
 Besides, $ \varphi, \varphi_1, \varphi_2 $  are STL formulas and 
 $\neg$ denotes negation,
 $\wedge$ and $\vee$ denote conjunction and disjunction respectively. Let 
$[a,b]$ denote a time interval,  $\Diamond_{[a,b]}(\varphi)$ is the future operator, meaning that $\varphi$ should hold at  some time in $[a,b]$,
$\square_{[a,b]}(\varphi)$ is the globally operator, meaning that $\varphi$ should hold throughout the interval $[a,b]$. Besides,   $\rightarrow$ denotes implication. 
One can generate an arbitrarily complex STL formula by recursively applying STL operations.
For example, $\square_{[a_1,b_1]}( \Diamond_{[a_2,b_2]}(\varphi)  )$. In particular, $( {\xi}, t) \models \varphi$ denotes the satisfaction relation and means that the signal $x$ satisfies the STL formula $\varphi$ at time $t$.

\textbf{Quantitative Semantics of STL}:
The quantitative semantic of STL is defined to evaluate the satisfaction degree of STL formulas over time, with each operator offering a distinct interpretation of robustness. 
The quantitative semantics of some common STL operators are concluded in TABLE~\ref{tab:stl_operators} in Appendix,  and the semantics of more operators could be found in \citep{leung2023backpropagation, yu2024continuous}.  
Each operator in TABLE~\ref{tab:stl_operators} has a specific role in computing the satisfaction degree of an STL formula. 
Based on such quantitative semantic $\rho(\cdot)$, STL  can be integrated into optimization-based trajectory planning \citep{zhao2024survey},  
even gradient-based trajectory planning \citep{leung2023backpropagation} or control synthesis  \citep{yu2024continuous}. 
In this paper, DMPs will be modified and generalized within the optimization-based planning framework by the aid of 
open-source toolbox PyTeLo  \citep{cardona2023flexible}.

\begin{table}[!htb]
    \centering
    \begin{tabular}{|c|}
    \hline
      Quantitative semantics of basic STL operators \\
    \hline
     \(\rho(-\varphi,{\xi}, t)=-\rho(\varphi,{\xi}, t)\)   \\
      \( 
     \rho(\varphi_1\land \varphi_2 ,{\xi}, t)=\min \{\rho(\varphi_1,{\xi}, t), \rho(\varphi_1, {\xi}, t)\} \)   \\
      \( 
     \rho(\varphi_1\lor  \varphi_2 ,{\xi}, t)=\max \{\rho(\varphi_1,{\xi}, t), \rho(\varphi_1, {\xi}, t)\} \) \\
      \( 
     \rho(\square_{[a,b]}(\varphi) ,{\xi}, t)=\min_{t'\in[t+a,t+b]}~\rho(\varphi, {\xi}, t')\)   \\
    \(  \rho(\Diamond_{[a,b]}(\varphi) ,{\xi}, t)=\max_{t'\in[t+a,t+b]}~\rho(\varphi, {\xi}, t')\)   \\
      \(\cdots\)  \\
    \hline 
    \end{tabular}
    \caption{Quantitative semantics of basic STL operators. 
        }
    \label{tab:stl_operators}
\end{table}

\section{Problem Definition} \label{section:PD}

This section formally defines the problem of \textbf{long-horizon manipulation tasks} within the framework of \textbf{LfD} and \textbf{TAMP}.

We assume the availability of a {demonstration dataset} $\mathcal{D}$, containing multiple demonstration examples of various manipulation skills:
$ \mathcal{D} = \{ d_i^j \mid i = 1, \dots, \mathcal{I}, \ j = 1, \dots, \mathcal{J}(i)\} $.
Each demonstration $d_i^j = \{ \mathbf{p}_{t} \mid t = 1, \dots, T_i^j \}$ is a sequence of end-effector poses, representing the $j$-th demonstration of the $i$-th manipulation type. 
LfD methods learn and represent each kind of robot manipulation skill from each sub-group $\{d_i^j, \quad j = 1,...,\mathcal{J}(i) \}$ in the dataset $\mathcal{D}$.

The state of the robot $\mathbf{s} \in \mathbb{R}^n$, 
evolving over time $t$, with the initial state written as $s_0$, and the action of the robot can be written as $a_t$. It is remarkable that, with the direct pose controller of the robot end, the robot action can be represented as the command of the end effector pose: $a_t \triangleq \mathbf{p}_t \in \mathbb{R}^3 \times \mathbb{S}^3$.

To formally describe complex long-horizon tasks with temporal constraints, a rich and expressive language is required in advance.
We assume that a Signal Temporal Logic specification is given for this purpose, and the sequence of sub-tasks is indexed by $\epsilon \in \{ 1, \dots, N_\epsilon \}$ at the task level.

\subsubsection{Formal Problem Statement}
Given the following available information: 
\begin{enumerate}[label=\alph*.]
     \item A demonstration dataset $\mathcal{D}$ containing demonstrations of manipulation ~ actions.
     \item An STL specification $\Psi^{\text{STL}}_{\text{L-H task}}$ that describes the information of the long-horizon task.
     \item An initial robot state $s_0$.
\end{enumerate}

\textbf{The goal} is to generate a sequence of {actions} $\{ a_t \mid t = 1, \dots, T \}$, which corresponds to a sequence of end-effector poses trajectory $\{ \mathbf{p}(t) \mid t = 1, \dots, T \}$ when using the end effector pose controller of the robot.

It is worth mentioning that the generated sequence should be not only learned from demonstrations $\mathcal{D}$ (because the learned dynamic characteristics enhance the validity of the manipulations), but also follow the requirements of task and motion constrains defined by $\Psi^{\text{STL}}_{\text{L-H task}}$.

The above goal can be  divided into  two key sub-problems:
\begin{itemize} 
    \item \textbf{Task Planning:} Determining the correct sequence and specific structure of sub-tasks based on $\Psi^{\text{STL}}_{\text{L-H task}}$.
    \item \textbf{Motion Planning \& Generalization:} Generating and adjusting the specific motion trajectory for each subtask, such that it satisfies the overall task and motion satisfaction, while the dynamic characteristics are preserved.
\end{itemize}

\subsubsection{Proposed Approach Overview}

To address this problem, the proposed framework leverages the given hierarchical STL specification $\Psi^{\text{STL}}_{\text{L-H task}}$, as shown in Section \ref{sec:representSTL}. 
As detailed in Section \ref{sec:STL2BTs}, a portion of the STL specification is transformed into Linear Temporal Logic (LTL) to facilitate high-level task planning, resulting in a Behavior Tree that orchestrates the sequence of sub-tasks. 
Concurrently, as described in Section \ref{subsec:3.3}, the relevant STL constraints are directly passed to the motion planning layer. 
These constraints guide an optimization-based approach to adapt the learned DMPs, ensuring that the generated motions are both contextually appropriate and compliant with the specifications, thereby enabling the robot to successfully complete challenging long-horizon manipulations.

\section{Main Framework}
\label{sec:FrameworkLfD}
To address the problem defined in Section \ref{section:PD},
the objective in this section is to utilize and generalize LfD methods to long-horizon manipulation tasks, with specific task and motion requirements. Specifically, a hierarchical TAMP framework is proposed    
to ensure both effective decision-making and the generalization ability of learned skills across multi-stage scenarios during long-term operations, shown in Fig.~\ref{fig0}.

\subsection{STL Representation for Long-horizon Tasks} \label{sec:representSTL}

STL is a standard and useful way to represent the specified needs of different long-horizon tasks. The specific definition of STL is given in Section \ref{PRE}. Notably, several methods have been proposed to acquire the correct STL formula gained from human natural languages \cite{mao2024nl2stl, choe2025seeing, fang2025enhancing}. 
In this paper, it is assumed that the given STL $\Psi^\text{STL}_\text{L-H task}$ can describe a certain long-horizon task correctly. A valid representation of long-horizon tasks by STL is provided in this subsection.

In detail, the long-horizon task  $\Psi_\text{L-H task}^\text{STL}$  is defined recursively as follows: 
\begin{equation}
    \Psi^{\text{STL}}_\text{L-H task} = \Diamond_{(0, T_1)} \Psi_{{\epsilon} = 1}^{\text{STL}} \,\wedge  ... \,\wedge \Diamond_{(T_{N_\epsilon}, T)} \Psi_{{\epsilon = N_\epsilon}}^{\text{STL}} .
    \label{Psi_LHtask}
\end{equation}
Each sub-task STL $\Psi_{{\epsilon} }^{\text{STL}}$ represents the information of each individual sub-task. 
Taking the ``pick-move-place" task as an example:
\begin{equation}
    \Psi^{\text{STL}} = \Diamond_{(t_0, t_1)} \varphi_{\text{pick}}^{\text{STL}} \,\wedge \Diamond_{(t_1, t_2)} \varphi_{\text{move}}^{\text{STL}} \,\wedge \Diamond_{(t_1, t_2)} \varphi_{\text{place}}^{\text{STL}}, 
    \label{Psi_pmp}
\end{equation}
there are three atomic manipulations: $\varphi_{\text{pick}}^{\text{STL}}$, $\varphi_{\text{move}}^{\text{STL}}$ and $\varphi_{\text{place}}^{\text{STL}}$.
Further, for each of the atomic manipulations, $\psi^{\text{STL}}$ can be written as with pre-condition, post-condition, action and STL-constrains:
\begin{equation}
    \varphi^{\text{STL}} = \text{C}_{\text{post}} \,\vee \left[\text{C}_{\text{pre}} \,\wedge (\text{Action}^{\text{STL}} \,\wedge \text{C}_{\text{STL}})\right] .
    \label{stl3}
\end{equation}
For example,  the  atomic manipulations (move) $\varphi_\text{move}^{\text{STL}}$ can be specifically given by the following components:
\begin{equation}
    \begin{aligned}
    & \text{C}_{\text{post}}: \square_{[t_1, t_1 + \delta t]} \left(\left\| \mathbf{p}_{\text{cup}} - \mathbf{p}_{\text{c\_goal}} \right\|_2 < 0.01\right), \\[1ex]
    & \text{C}_{\text{pre}}:  \Diamond_{[0, t_1]} \varphi_{\text{pick}}^{\text{STL}}, \\[1ex]
    & \text{Action}^{\text{STL}}:\square_{[t_1, t_2]} \left(\left\| \mathbf{p}_{\text{cup}} - \mathbf{p}_{\text{DMP}}(t) \right\|_2 < 0.01\right), \\[1ex]
    & \text{C}_{\text{STL}}: \!  \square_{[t_1, t_2]} (\dot{\mathbf{y}} < 0.1) \wedge \square_{[t_1, t_2]} \left(\left\| \mathbf{p}_{\text{cup}} - \mathbf{p}_{\text{obs}} \right\|_2 > 0.1\right). \\[1ex]
    \end{aligned}
    \label{stl4}
\end{equation}
In \eqref{stl4}, the post-condition $\text{C}_{\text{post}}$ means that the cup is required to be placed on the right place after the success of the move action. The pre-condition $\text{C}_{\text{pre}}$ denotes that the ready-to-move cup has been picked already, which can be indicated by the last action - pick action state. The Action part - $\text{Action}^{\text{STL}}$ requires that the robot end effector should follow the trajectory generated by motion planner  (DMP). The STL constraint $\text{C}_{\text{STL}}$ represents requirements of safe velocity and obstacle avoiding. 

\begin{figure}[!h]
        \centering
        \includegraphics[width=0.75\hsize]{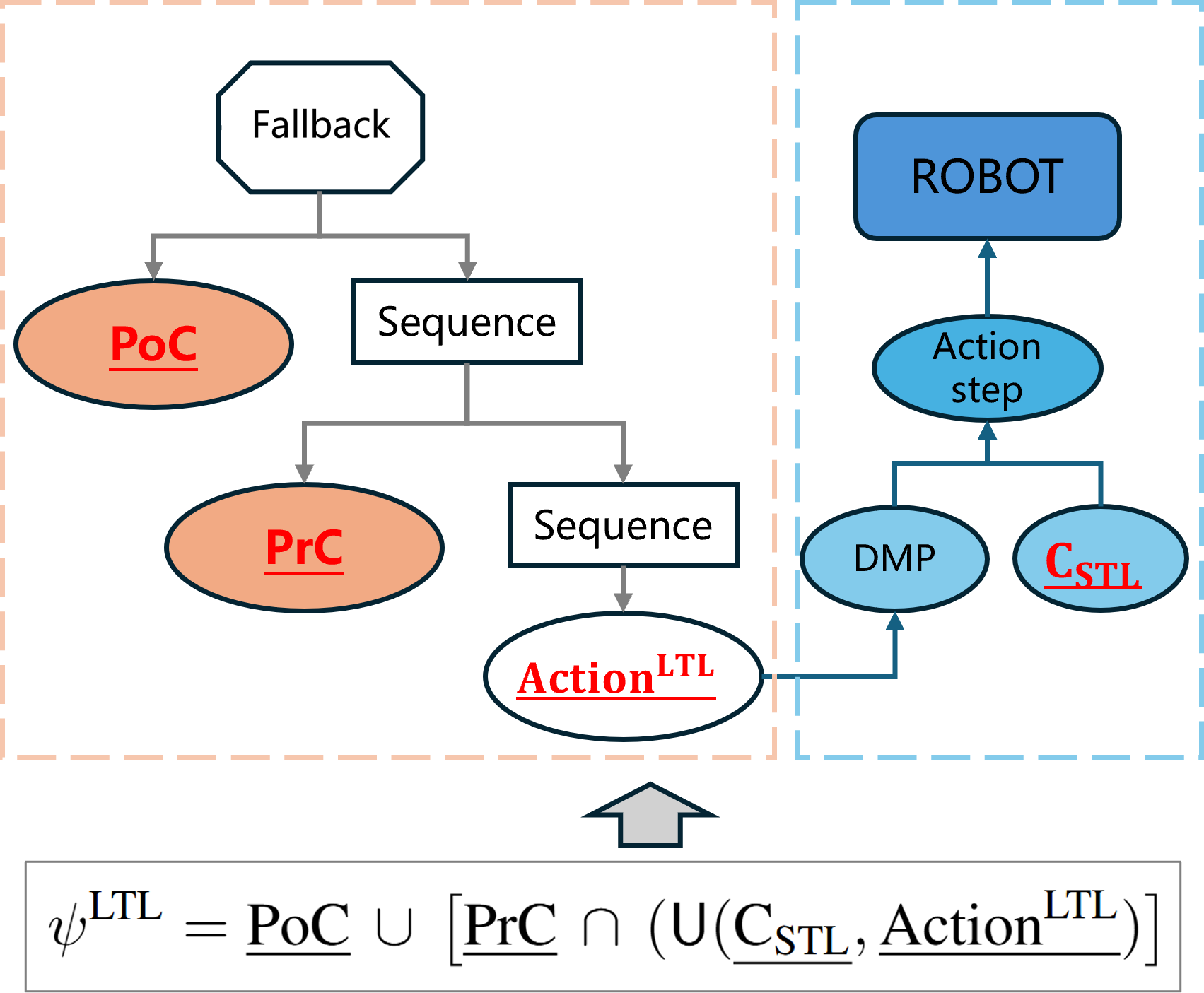}
        \caption{The utilization of LTL and STL to generate BT (in task planning) and constrain representation (in motion planning). }
        \label{geneBTs}
\end{figure}

Within the above part,  an example is shown to illustrate how to express the long-horizon tasks by STL.  
Motivated by the idea of designing behavior trees from goal-oriented LTL formulas \cite{neupane2023designing} and the idea of 
translating SpaTiaL (a STL-like specification) into an LTL formula \cite{pek2023spatial}, 
the above STL representation (except for  $\text{C}_{\text{STL}}$) will be transformed into LTL-like formulas to build BTs for task planning in Section \ref{sec:STL2BTs}. And $\text{C}_{\text{STL}}$ will be applied for motion planning in Section \ref{subsec:3.3}.
The utilization of the above temporal logics is shown in Fig.~\ref{geneBTs}.

\subsection{Task planner with Generated BTs}\label{sec:STL2BTs}

As stated in the last paragraph of Section \ref{sec:representSTL}, to generate BTs,  the STL representation will be transformed into LTL formulas by abstracting spatial relations into atomic propositions.

Firstly, the LTL formula is introduced here.
In detail, $\mathsf{F}( \psi)$ (“eventually $\psi$”) holds if there exists some future time point at which the formula $\psi$ is true,
$\mathsf{G}( \psi)$ (“globally $\psi$” or “always $\psi$”) holds if $\psi$ is true at the current time and at every future time point, and
$\mathsf{U}(\psi_1, \psi_2)$ (“$\psi_1$ until $\psi_2$”) holds if there is some future time point at which $\psi_2$ becomes true, and at all time points prior to that one, $\psi_1$ remains true. By the aid of these notations, 
 the STL formulas in \eqref{Psi_pmp}, \eqref{stl3} and \eqref{stl4} can be abstracted as LTL formulas respectively:
\begin{align}
    \Psi^\text{LTL} &= \mathsf{U} (\mathsf{F}(\psi_\text{Pick}^{\text{LTL}}, \mathsf{U} (\mathsf{F} \psi_\text{move}^{\text{LTL}},\mathsf{F}\psi_\text{place}^{\text{LTL}}))),
    \label{eq_pmp} \\
  \psi^{\text{LTL}} &= \text{PoC} \,\vee \left[\text{PrC} \,\wedge ( \mathsf{U} (\text{C}_{\text{STL}}, \text{Action}^{\text{LTL}} )\right] ,
    \label{eq:philtl}
\end{align}
and
\begin{equation}
\begin{aligned}
    & \text{PoC}: CupGoal(\mathbf{p}_\text{goal}), \\[1ex]
    & \text{PrC}:  Picked, \\[1ex]
    & \text{Action}^{\text{LTL}}:{DMP}_\text{move}. \\[1ex]
\end{aligned}
\end{equation}
Based on this,  behavior trees could be 
designed and built automatically form LTL formulas as in  \cite{neupane2023designing}.
As shown in Fig.~\ref{geneBTs}, 
the LTL representations are used to generate the BTs, following the thought in \cite{neupane2023designing}, while the STL part $\text{C}_\text{STL}$ is passed to the motion planning module, which will be illustrated in Section \ref{subsec:3.3}.


\begin{figure*}[!hbt]
        \centering
        \includegraphics[width=0.88\hsize]{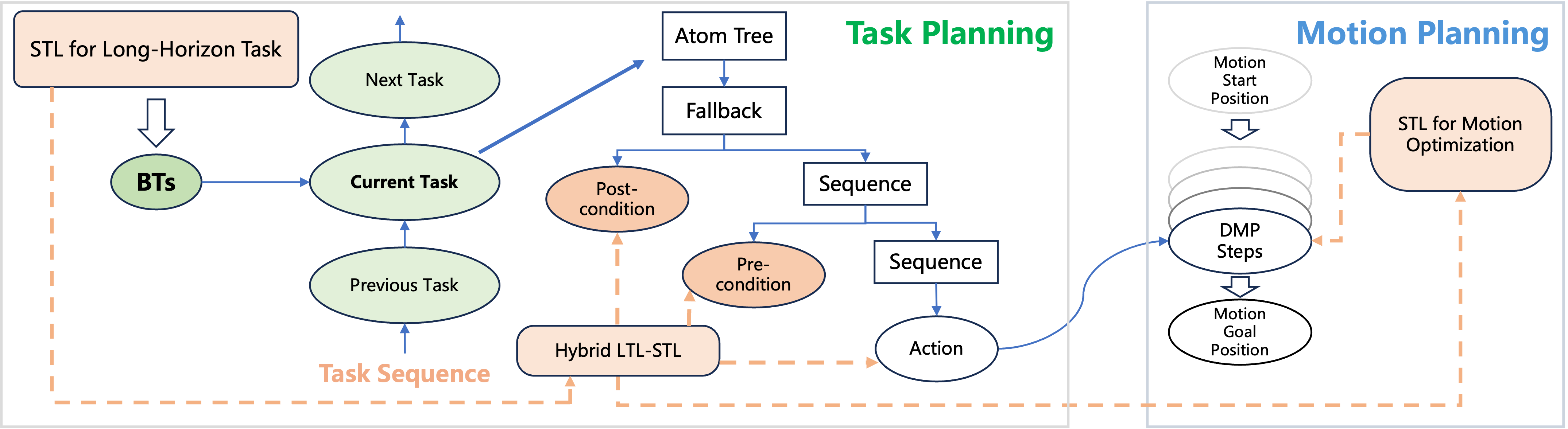}
        \caption{The overall task and motion level planning framework.}
        \label{fig0}
\end{figure*}

Then, the constructed BTs are then used to coordinate the execution of manipulation skills by encoding each action node with associated pre- and post-conditions.

In the remained context of the section, the role of the constructed BT is discussed. 
Suppose that the $\epsilon$-subtask is executed using a DMP. This DMP is learned from demonstrations $\mathcal{D}$ via GMM-GMR (as will be shown in Section~\ref{subsec:3.3}) and can be represented in a DS-based formulation:

\begin{equation}
 \ddot{\mathbf{p}} = f_{\epsilon}(\dot{\mathbf{p}}, \mathbf{p}, \mathbf{p}_{\text{init}}^{\epsilon}, \mathbf{p}_{\text{goal}}^{\epsilon}, \mathbf{F}_{\text{lrn}}^{\epsilon} \mid \mathcal{D}),
 \label{eq:LfDp}
\end{equation}
where $\mathbf{p} = (\mathbf{y}, \mathbf{q}) \in \mathbb{R}^3 \times \mathbb{S}^3$ and $f_{\epsilon}$ denotes the DMP dynamics for each subtask $\epsilon$. The notations $\mathbf{p}_\text{init}$ and $\mathbf{p}_\text{goal}$ are the parameters of the DMPs, while $\mathbf{F_\text{lrn}^{\epsilon}}$ represents the forcing term learned from demonstrations. From the control–theoretic viewpoint
in Section \ref{sec:BT}, 
the constructed BTs are applied to generate the switching signal $\epsilon(\mathbf{s})$ based on the system state $\mathbf{s}$, enabling a switching control strategy:
\begin{equation}
    \ddot{\mathbf{p}} = f_{\epsilon(\mathbf{s})}(\dot{\mathbf{p}}, \mathbf{p}, \mathbf{p}_{\text{init}}^{\epsilon}, \mathbf{p}_{\text{goal}}^{\epsilon}, \mathbf{F}_{\text{lrn}}^{\epsilon} \mid \mathcal{D}).
    \label{eq:bt_switch}
\end{equation}

\subsection{Motion Planner with STL-DMPs}\label{subsec:3.3}

According to the above subsections, the architecture of BTs provides an effective task-level planner with the reactivated ability when unexpected disturbances occur. 
However,  how to enable generalization across task instances, as well as additional constraints $\varphi =\text{C}_{\text{STL}}$ in new scenarios are not discussed. 
This section shows an efficient way to incorporate  $ \text{C}_{\text{STL}}$ into DMPs by optimization-based trajectory planning.

More specifically, based on the background  about STL in Section \ref{sec:STL}, the quantification of STL fragment $\text{C}_\text{STL}$ in \eqref{stl4} is used to optimize the learned DMPs, where the dynamic characteristics \textbf{learned from demonstrations} are retained and \textbf{generalized} to new environments, as  shown in  Fig.~\ref{opti_fig}. 

For brevity, the learned parameters $\mathbf{F_\text{lrn}^{\epsilon} | \mathcal{D}}$ is abbreviated as $\mathbf{F_\text{lrn}^{\epsilon}}$. 
To preserve the learned dynamic from demonstration while satisfying STL constraint  $\varphi = \text{C}_\text{STL}$, we formulate an optimization-based method to modify the {original} \textit{learned} parameter vector  $\mathbf{F_\text{lrn}^{\epsilon}}$ to another \textit{optimized} parameter vector $\mathbf{F_\text{opt}^{\epsilon}}$. 
With the proposed optimization mechanism, the constructed BT-triggered  DMPs in \eqref{eq:bt_switch} can be modified into:
\begin{equation}\label{eq:pF}
    \ddot{\mathbf{p}}   =   f_{\epsilon(\mathbf{s})}(
 \dot{\mathbf{p}}, {\mathbf{p}},  \mathbf{p}_{\text{init}}^{\epsilon}, \mathbf{p}_{\text{goal}}^{\epsilon}, \mathbf{F_\text{opt}^{\epsilon}}),
\end{equation}
where the forcing terms of DMPs are optimized from $\mathbf{F}_\text{lrn}^{\epsilon}$ to $\mathbf{F}_\text{opt}^{\epsilon}$ to generalize in new scenarios.

The following context of this section introduces the detailed procedure of modifying parameters of DMPs by the STL formula $\varphi$ in a forcing-term-based optimization.

\subsubsection{Proposed Optimization Function for DMPs}

First, for each motion planning process in sub-task~$\epsilon$, the forcing term~$\mathbf{F}^{\epsilon}$ is represented as a time-indexed vector of length $T$, with either 3 dimensions $\mathbb{R}^{T \times 3}$ (for position in~$\mathbb{R}^3$) or 7 dimensions $\mathbb{R}^{T \times 7}$ (for pose in~$\mathbb{R}^3 \times \mathbb{S}^3 \subset \mathbb{R}^3 \times \mathbb{R}^4$).
Then the following objective function  quantifies the difference between the optimized and original DMP forcing terms: 
\begin{equation}
J^{0}_{\text{DMP}}(\mathbf{F}^{\epsilon}) = \| \mathbf{F}^{\epsilon} - \mathbf{F}_{\text{lrn}}^{\epsilon} \|_2, 
\label{eq:dmp_objective_2}
\end{equation}
which can better preserve the effective dynamic features in human demonstrations than the common trajectory-based optimization function:
$J'_{\text{DMP}}(\mathbf{p}^{\epsilon}) = \| \left\{ \mathbf{p}^{\epsilon} \right\} - \left\{  \mathbf{p}_{\text{lrn}}^{\epsilon}\right\} \|_2$. 
The objective function \eqref{eq:dmp_objective_2} offers distinct advantages over direct trajectory optimization approaches $J'_{\text{DMP}}(\mathbf{p}^{\epsilon})$ by inherently maintaining the natural force features and the convergence properties of DMPs to target points.  

 Moreover, to enhance generalization across multiple demonstrations, we utilize the means and variance learned by {Gaussian Mixture Model-Gaussian Mixture Regression} (GMM-GMR) when learning DMPs from multiple demonstrations, in Section \ref{subsec:LfD}. This probabilistic approach enables adaptive weighting of the forcing function optimization based on trajectory variance analysis. For clarity of the main text, the specific design of the weighting part $\mathcal{W}$ is shown in Appendix \ref{sec:OFC}. 
 With the weighting matrix $\mathcal{W}$  learned from demonstrations, the DMPs optimization function $J^{0}_\text{DMP}$ in \eqref{eq:dmp_objective_2} can be enhanced  as:
\begin{equation}
J_{\text{DMP}}(\mathbf{F}^{\epsilon})  = \| \mathcal{W} 
\mathbf{F}^{\epsilon} -\mathcal{W} \mathbf{F}_{\text{lrn}}^{\epsilon} \|_2
\label{eq:dmp_objective}.
\end{equation}

\subsubsection{Optimization with Both DMP and STL Objective Functions}
 Apart from the DMPs objective function in \eqref{eq:dmp_objective}, 
the STL objective function $J_{\text{STL}}(\mathbf{F^{\epsilon}})$ is defined by 
\begin{equation}\label{eq:Jstl}
    J_{\text{STL}}(\mathbf{F^{\epsilon}}) =   \rho(\varphi 
 ,\mathbf{p}, 0)  , 
\end{equation}
where the value $\rho(\varphi,\mathbf{p}, 0)$ denotes the quantitative semantics of STL specification  $\varphi$ for signal $\mathbf{p}$ at time zero (as given in Section \ref{sec:STL}).
In \eqref{eq:Jstl}, the STL specification $\varphi$ describes the constraints for task demands and the motion limitations, and the signal $\mathbf{p}$ is the desired pose (action) given by the DMP with parameters $\mathbf{F^{\epsilon}}$:
\begin{align}
 \ddot{\mathbf{p}}   &= 
 f_{\epsilon}(
 \dot{\mathbf{p}}, {\mathbf{p}},  \mathbf{p}_{\text{init}}^{\epsilon}, \mathbf{p}_{\text{goal}}^{\epsilon},  \mathbf{F^{\epsilon}}). \label{eq:LfDp}
\end{align}

\begin{figure}[!t]
        \centering
        \includegraphics[width=0.9\hsize]{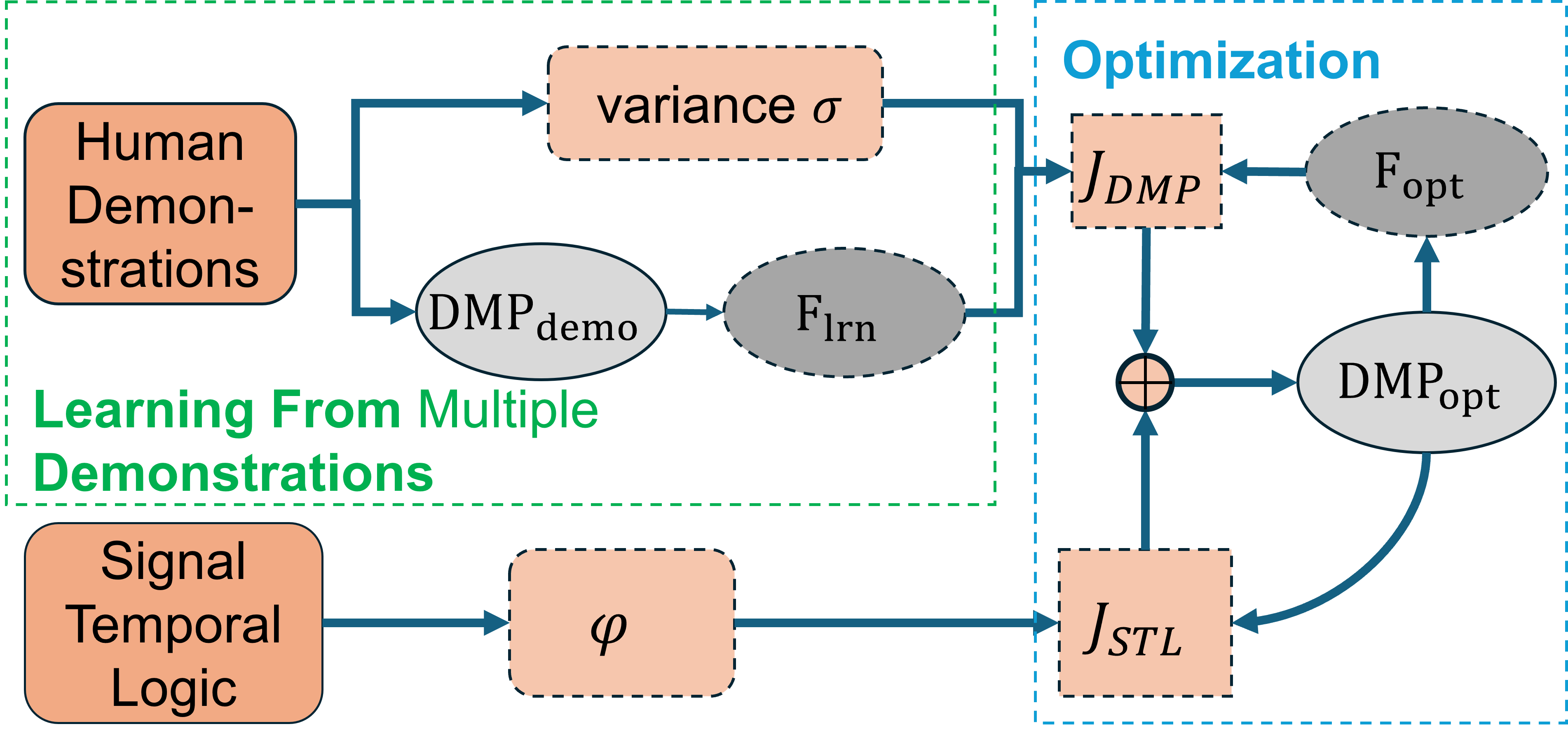}
        \caption{Motion planning framework: the proposed method generalize the DMPs trajectory with extensive STL-represented information, while preserve the nature dynamic learned from multi-demonstrations.}
        \label{opti_fig}
\end{figure}

Based on the above two facts, the parameter vector 
$\mathbf{F_\text{opt}^{\epsilon}}$ required in \eqref{eq:pF} is determined  by the following optimization: 
\begin{equation}\label{eq_opt}
\mathbf{F^{\epsilon}_{\text{opt}}}=
\underset{\mathbf{F^{\epsilon}}}{\operatorname{argmin}}
  ~~ \lambda_1 J_{\text{STL}}(\mathbf{F^{\epsilon}}) + \lambda_2 J_{\text{DMP}}(\mathbf{F^{\epsilon}}) , 
\end{equation}
where \( \lambda_1, \lambda_2 \in \mathbb{R}^+ \) are regularization parameters.
The overall framework of this subsection is shown in Fig.~\ref{opti_fig}.
It is worth noticing that the procedure of calculating the value of 
$J_{\text{STL}}(\mathbf{F^{\epsilon}})$ can be obtained by the aid of the recent open-source toolbox PyTeLo  \citep{cardona2023flexible}.  

As a summary, the optimized DMPs in  \eqref{eq:pF} (with parameters $\mathbf{F^{\epsilon}_{\text{opt}}}$ given by \eqref{eq_opt}) could generate new trajectories that simultaneously satisfy temporal logic specifications while preserving the demonstrated dynamic characteristics.

Generally speaking, Section \ref{sec:FrameworkLfD} proposes a hierarchical framework integrating BT, STL, and DMPs to augment LfD for long-horizon tasks in a TAMP framework (shown in Fig.~\ref{fig0}). The general STL representations are transformed into LTL form, which is then utilized to construct a behavior tree (task planner). By embedding STL constraints into atomic BT nodes, the framework ensures systematic propagation of spatiotemporal task-motion requirements. Moreover, an STL-constrained DMP optimization method preserves learned dynamics while adapting skills to new scenarios. The synergy between BT reactivity and STL formalization enables robust generalization from demonstrations to execution, bridging symbolic task planning with continuous motion control.


\begin{figure*}[!b]
    \centering
    \begin{subfigure}[t]{0.32\textwidth}
        \centering
        \includegraphics[width=\textwidth]{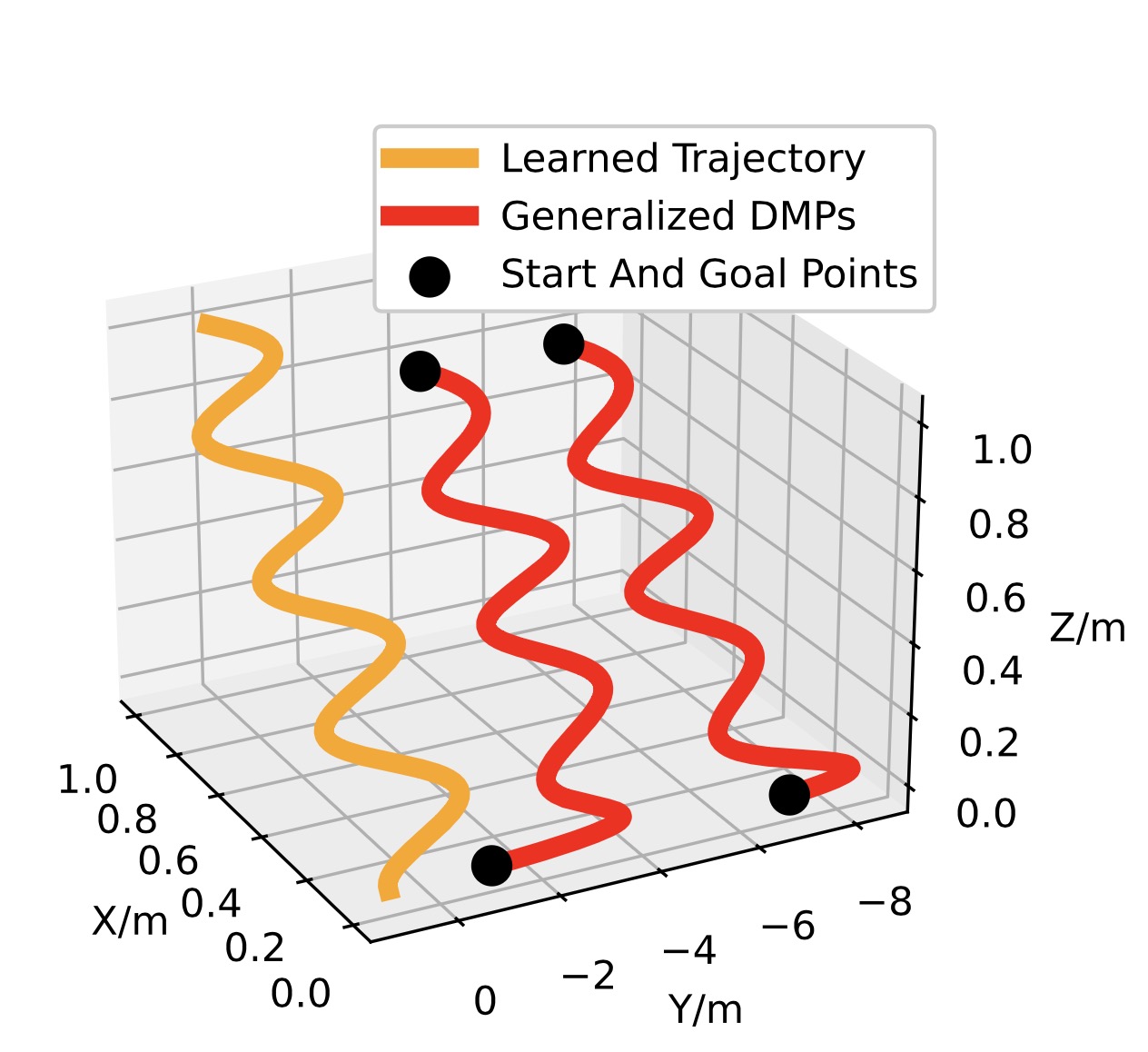}
        \caption{The performance of generalizing DMPs with new start and goal positions.}
        \label{simulation_DMP_base}
    \end{subfigure}
    \hfill
    \begin{subfigure}[t]{0.27\textwidth}
        \centering
        \includegraphics[width=\textwidth]{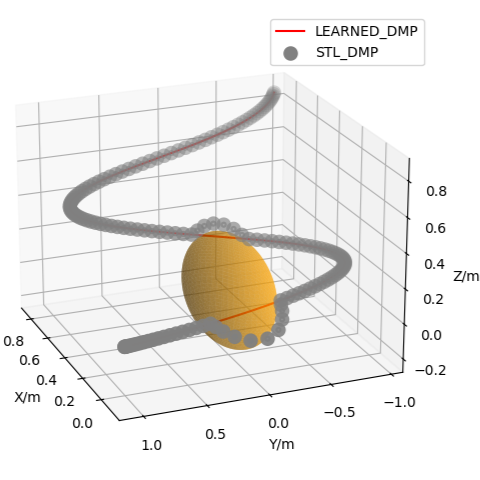}
        \caption{Unnatural obstacle avoidance generalizing performance with the trajectory-based  $J'_\text{DMP}$.}
        \label{simulation_compare_Y}
    \end{subfigure}
    \hfill
    \begin{subfigure}[t]{0.27\textwidth}
        \centering
        \includegraphics[width=\textwidth]{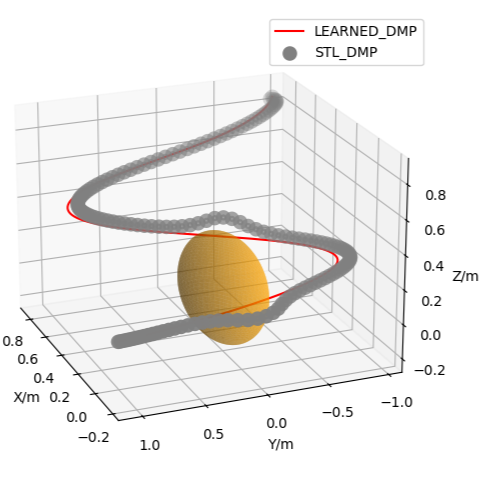}
        \caption{Excellent obstacle avoidance generalizing performance with the proposed forcing term-based $J_\text{DMP}$, in equation \eqref{eq:dmp_objective}.}
        \label{simulation_compare_F}
    \end{subfigure}
    \caption{Comparison of DMP simulations: 
      (a) generation with new start/goal positions, 
      (b) obstacle avoidance via trajectory-based objective, 
      (c) obstacle avoidance via forcing-term objective.}
    \label{fig:combined_DMP_simulations}
\end{figure*}

\section{Simulation}
This section presents a comprehensive evaluation of the motion planning part (illustrated in Section \ref{subsec:3.3}) in the proposed framework through systematic simulation. In the following part, we investigate the generalization capability of Dynamical Movement Primitives under task and motion constraints, which are encoded via Signal Temporal Logic, focusing on how spatiotemporal requirements are propagated to motion execution.

DMPs can learn the dynamical characteristic from the demonstration, and conveniently reproduce the learned skills with different start and goal positions. Moreover, it is important to guarantee the convergence of the goal for many kinds of robot manipulation skills. The reproduction ability is shown in Fig.~\ref{simulation_DMP_base}, where two trajectories of generated DMPs (represented by Red line) can generalize the demonstration trajectory (yellow line) into the new cases of different start and goal points, while presenting the dynamic nature.

Remaining the dynamical features is crucial for the validity when generalizing the learned skills, especially for continuous and high-dynamics manipulation skills, such as throwing objects and pouring water.
The comparison simulation results are shown in Fig.~\ref{fig:combined_DMP_simulations}, represented in an obstacle-avoiding scenario. The usual trajectory-based objective function, written as $J'_{\text{DMP}}(\mathbf{p}^{\epsilon}) = \| \left\{ \mathbf{p}^{\epsilon} \right\} - \left\{  \mathbf{p}_{\text{lrn}}^{\epsilon}\right\} \|_2$, is shown in Fig.~\ref{simulation_compare_Y} with an unnatural generalization. While the proposed forcing-term-based objective function, written as equation \eqref{eq:dmp_objective}, generating a more natural and a similar dynamical motion in  Fig.~\ref{simulation_compare_F}.
Using the difference of the forcing terms in section \ref{subsec:3.3} can better maintain the skill features than directly comparing the similarity of trajectories.

\begin{figure*}[!b]
    \centering
    \begin{subfigure}[t]{0.24\textwidth}
        \centering
        \includegraphics[width=\textwidth]{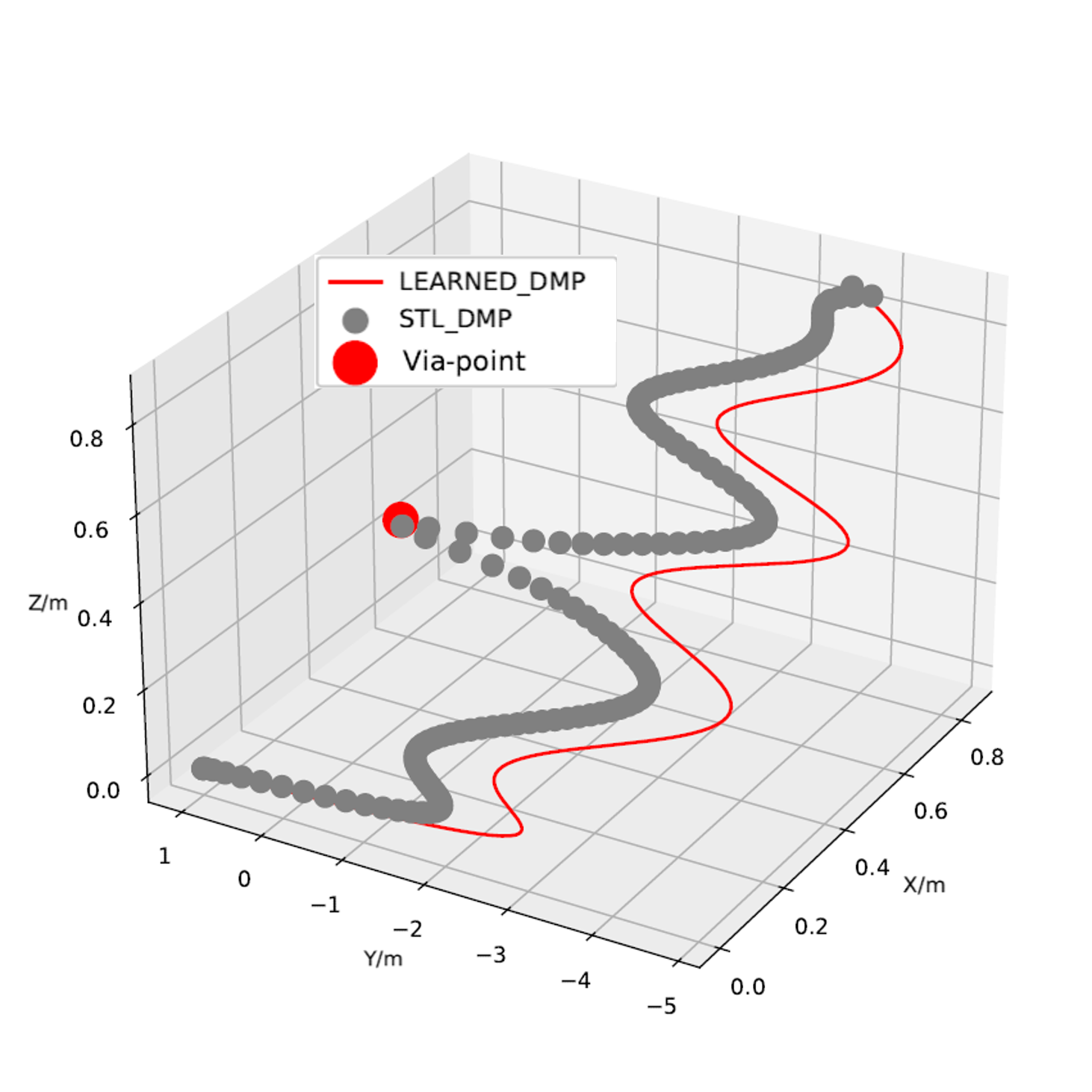}
        \caption{Via‐point scenario.}
        \label{simulation_via_point}
    \end{subfigure}
    \hfill
    \begin{subfigure}[t]{0.24\textwidth}
        \centering
        \includegraphics[width=\textwidth]{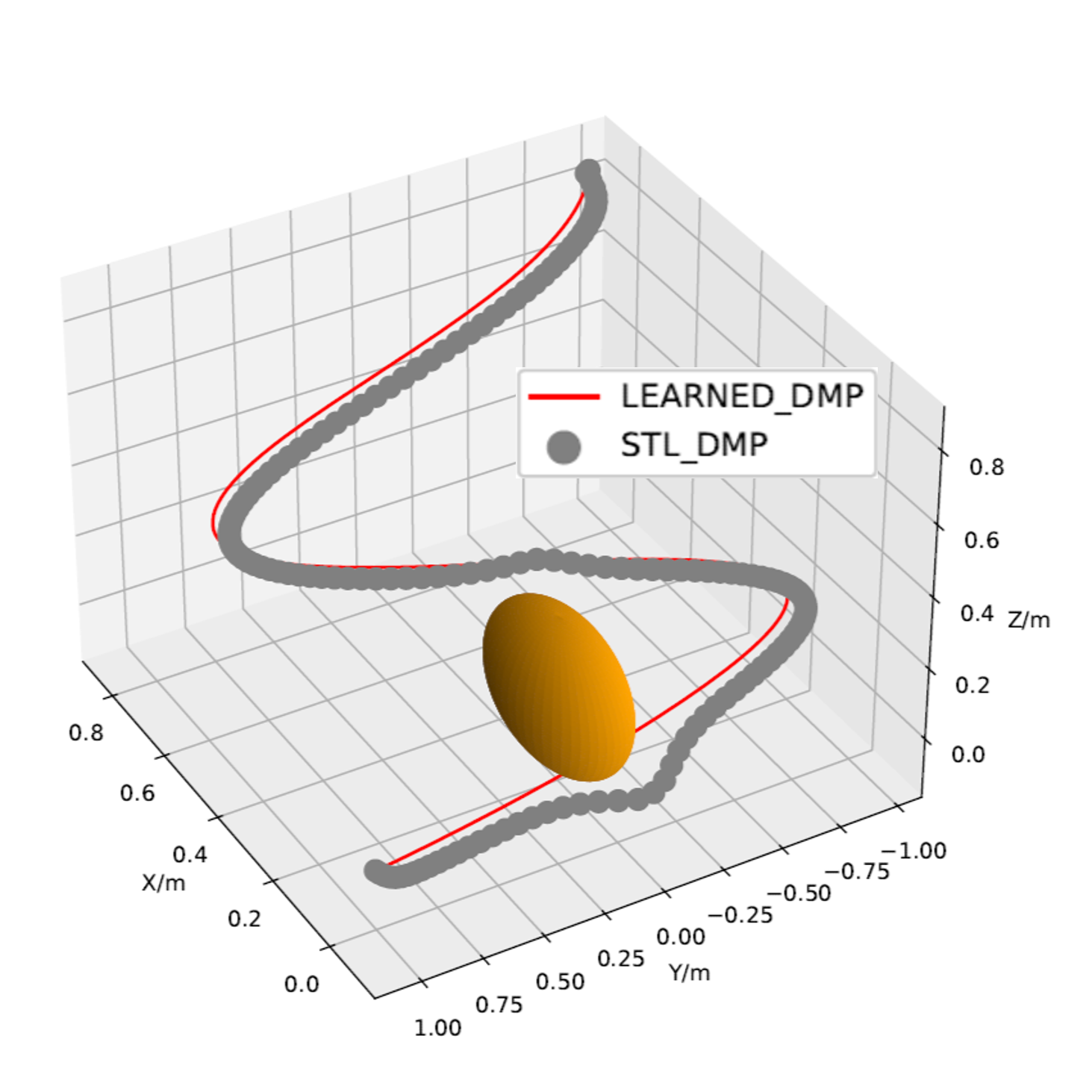}
        \caption{Obstacle scenario.}
        \label{simulation_obs_avoid}
    \end{subfigure}
    \hfill
    \begin{subfigure}[t]{0.24\textwidth}
        \centering
        \includegraphics[width=\textwidth]{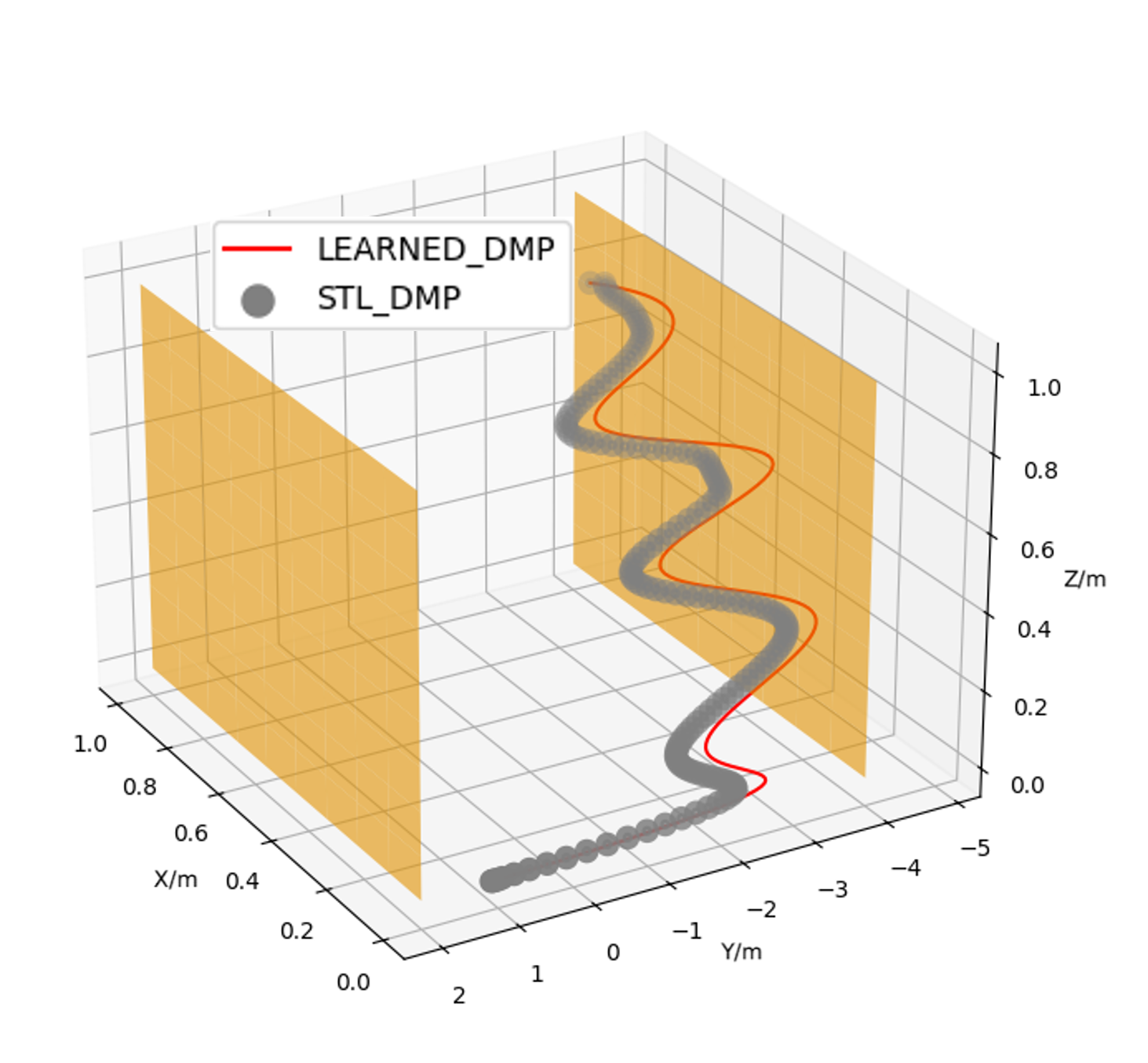}
        \caption{Space limit scenario.}
        \label{simulation_space_limit}
    \end{subfigure}
    \hfill
    \begin{subfigure}[t]{0.24\textwidth}
        \centering
        \includegraphics[width=\textwidth]{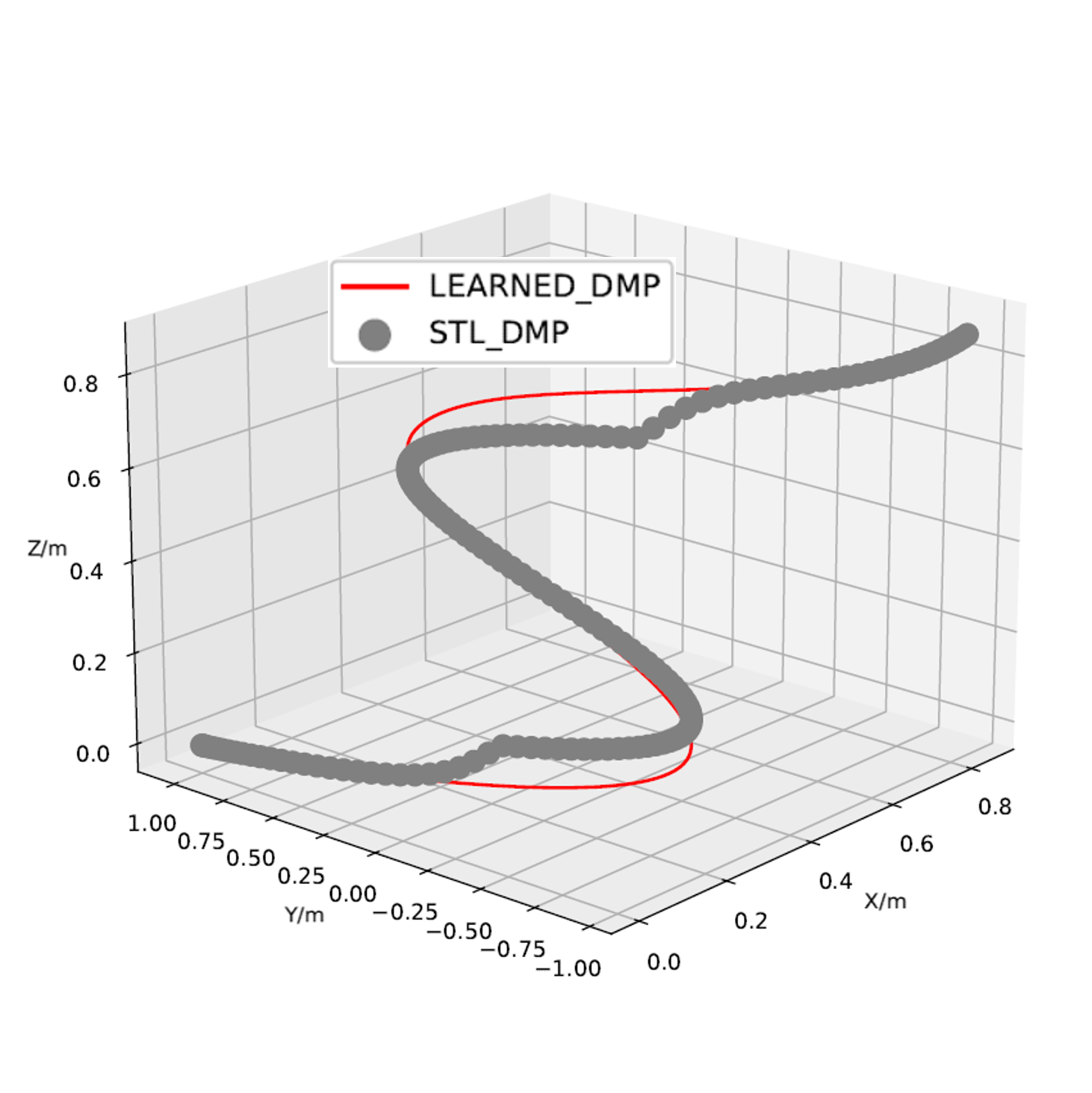}
        \caption{Velocity limit scenario.}
        \label{simulation_velocity_limit}
    \end{subfigure}

    \vspace{1em}

    \begin{subfigure}[t]{0.45\textwidth}
        \centering
        \includegraphics[width=\textwidth]{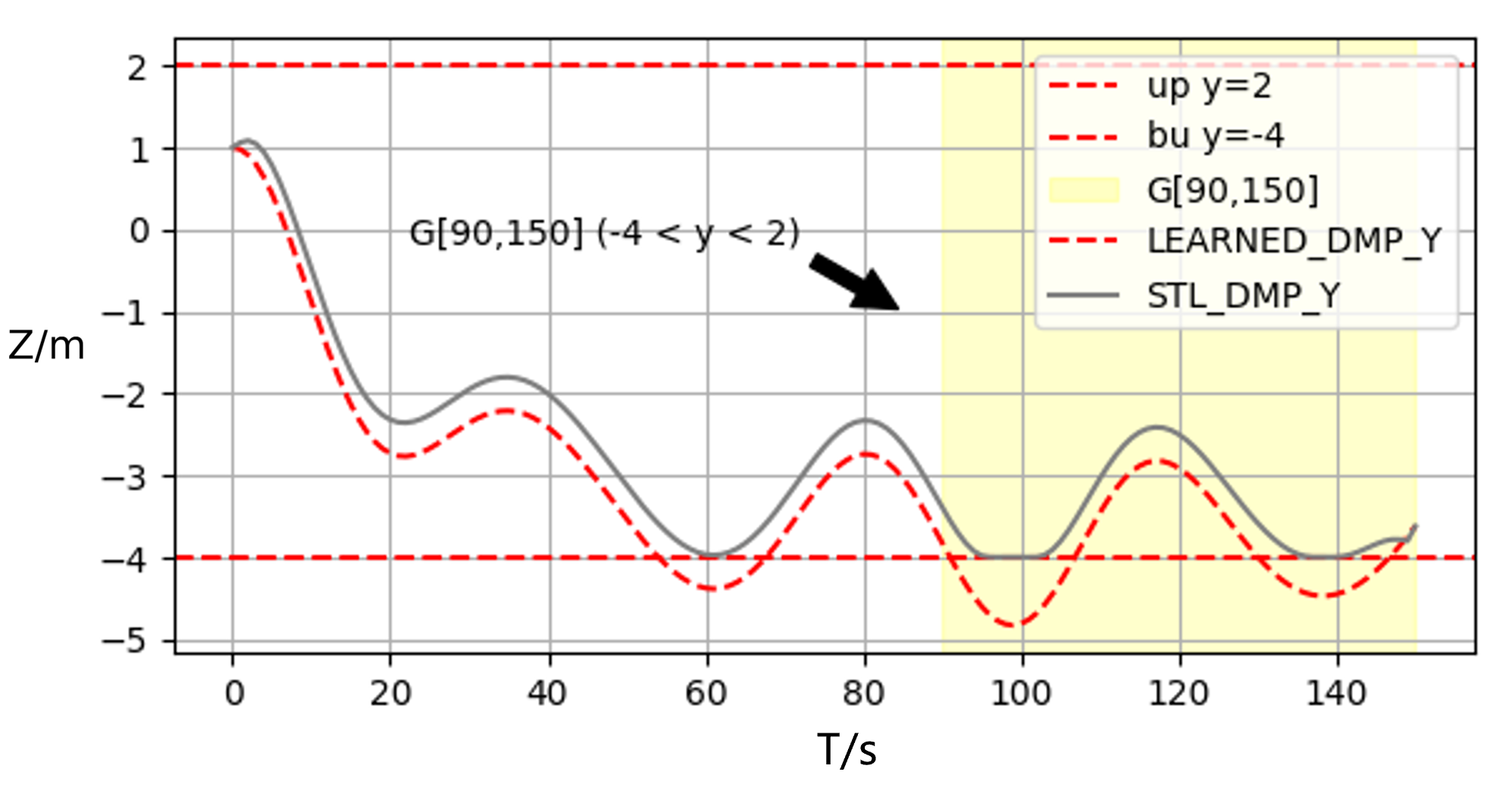}
        \caption{Temporal variation of $y$ under space limit.}
        \label{simulation_space_limit_2}
    \end{subfigure}
    \hfill
    \begin{subfigure}[t]{0.45\textwidth}
        \centering
        \includegraphics[width=\textwidth]{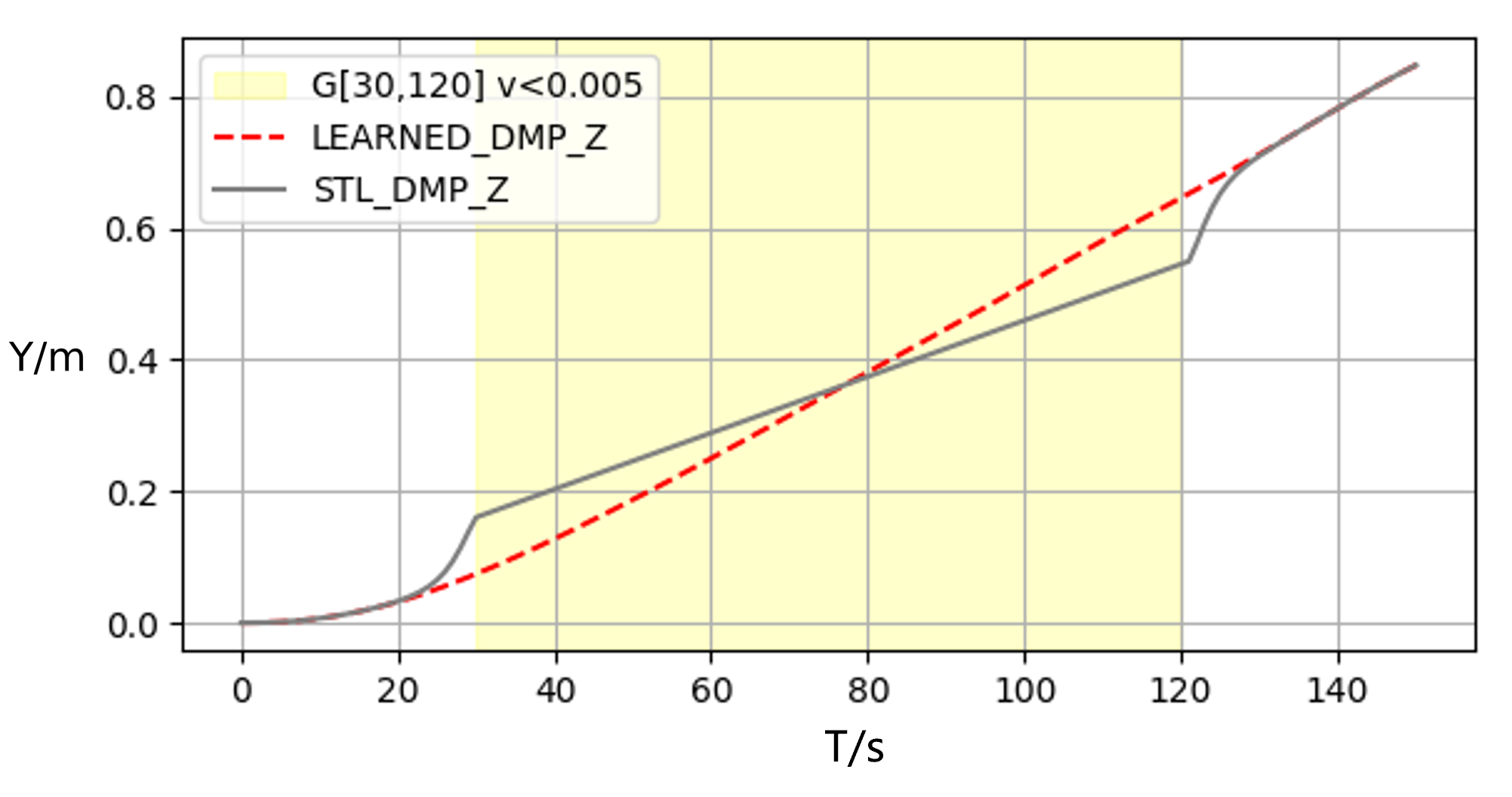}
        \caption{Temporal variation of $z$ under velocity limit.}
        \label{simulation_velocity_limit_2}
    \end{subfigure}

    \caption{Different scenarios for STL-DMP generalization performance: first row shows via-point, obstacle avoidance, space‐limit, and velocity‐limit scenarios performances, second row shows the corresponding state temporal variations in space‐limit and velocity‐limit scenarios. Each STL-DMP in figure represents the DMP generalized with the STL constraints in TABLE~\ref{table:simulation}, following the proposed optimization in Section \ref{subsec:3.3}.}
    \label{fig:rearranged_stl_dmp}
    
\end{figure*}

More simulation studies are employed to validate the effectiveness of the proposed methods in generalizing DMP-based manipulation skills under various spatio-temporal constraints. 
STLs can convey various task-level and motion-level demands for each task, generalizing DMPs in the motion planning part. Four representative scenarios: 
\textbf{via-point}, \textbf{obstacle avoidance}, \textbf{space limits}, and \textbf{velocity limits}, are shown in Fig.~\ref{simulation_via_point}, \ref{simulation_obs_avoid}, \ref{simulation_space_limit} and \ref{simulation_velocity_limit} respectively, the corresponding language descriptions, STL formulas and the practical usages are shown in TABLE~ \ref{table:simulation}. These examples indicate four kinds of task-level or motion-level constraints, which show the comprehensive application of the proposed method in extensive robot manipulation scenarios. 

\begin{table*}[!b]
\centering
\begin{tabular}{|c|c|c|}
\hline
Description &  STL formula & Feasible Usage \\
\hline
via-point &  \(\Diamond_{[0,150]}(||\mathbf{y}-\mathbf{y}_\text{via}||_2 < 0.01)  \) &  moving through critical inspection points  \\
obstacle avoidance &  \(\square_{[0,150]}(||\mathbf{y}-\mathbf{y}_\text{center}||_2 > R)  \) &  moving with unexpected obstacles \\
space limit & \(\square_{[90,150]}( -4 < y <2  )\) &  operating in confined laboratory spaces \\
velocity limit & \(\square_{[30,120]}(\dot{z} <0.005)\) &  carrying a steady-state object (a cup of tea) \\

\hline
\end{tabular}
\caption{The corresponding descriptions, STL formulas and the practical usages in Fig.~\ref{fig:rearranged_stl_dmp}.}
\label{table:simulation}
\end{table*}

\section{Experiments}\label{sec:exper}
To validate the proposed BT-TL-DMPs framework in real-world settings, we conduct two long-horizon robotic manipulation tasks. These experiments aim to demonstrate how the integration of Behavior Trees, Signal Temporal Logic, and Dynamical Movement Primitives enables modular, reactive, and generalizable robotic behaviors under complex spatio-temporal constraints. 
\subsection{Experiment Configuration}

\begin{figure*}[!htb]
    \centering
    \begin{subfigure}[b]{0.3\textwidth}
        \centering
        \includegraphics[width=\textwidth]{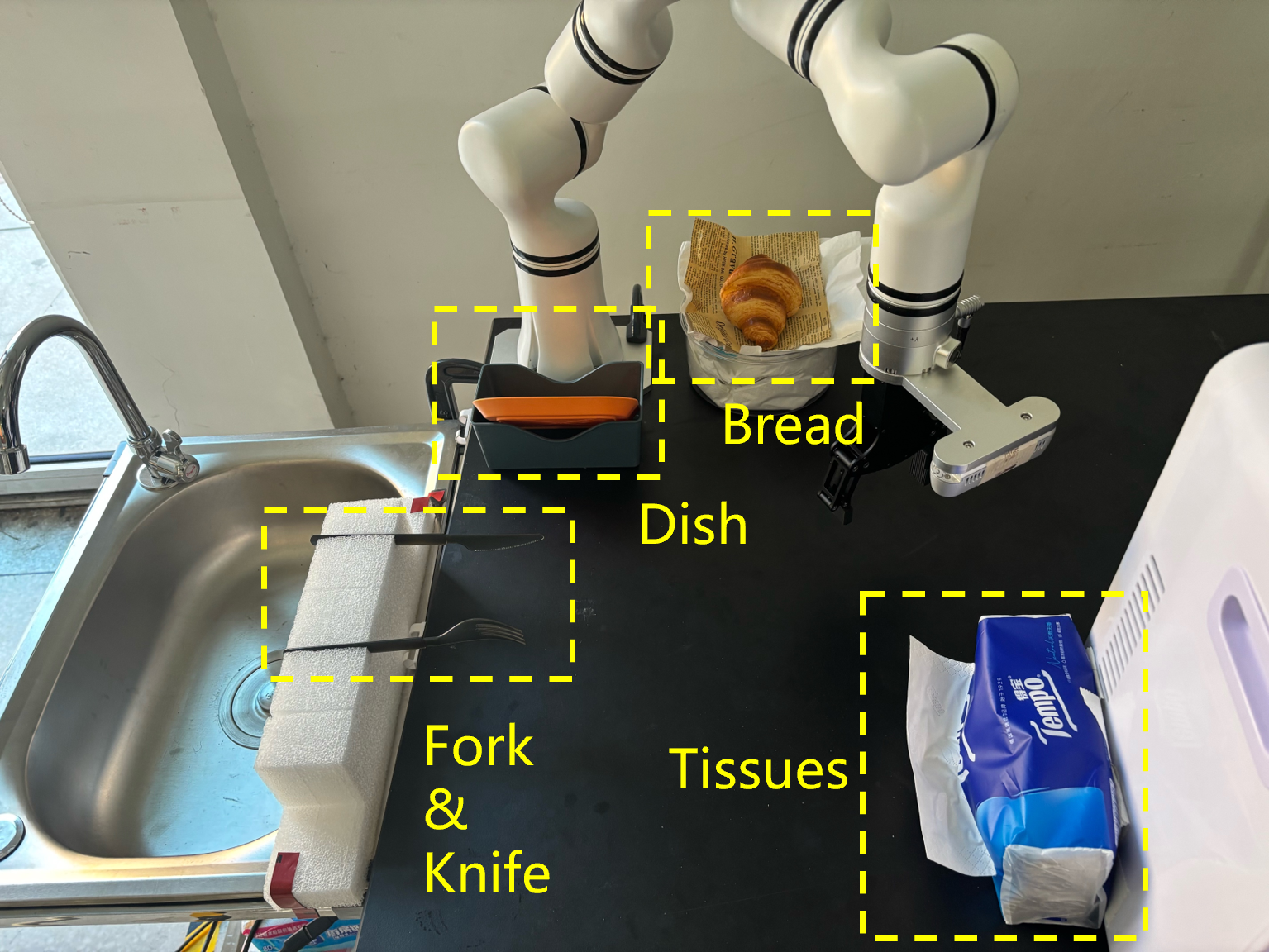}
        \caption{Scene before breakfast task.}
        \label{fig:exp_breakfast_1}
    \end{subfigure}
    \hspace{0.01\textwidth} 
    \begin{subfigure}[b]{0.3\textwidth}
        \centering
        \includegraphics[width=\textwidth]{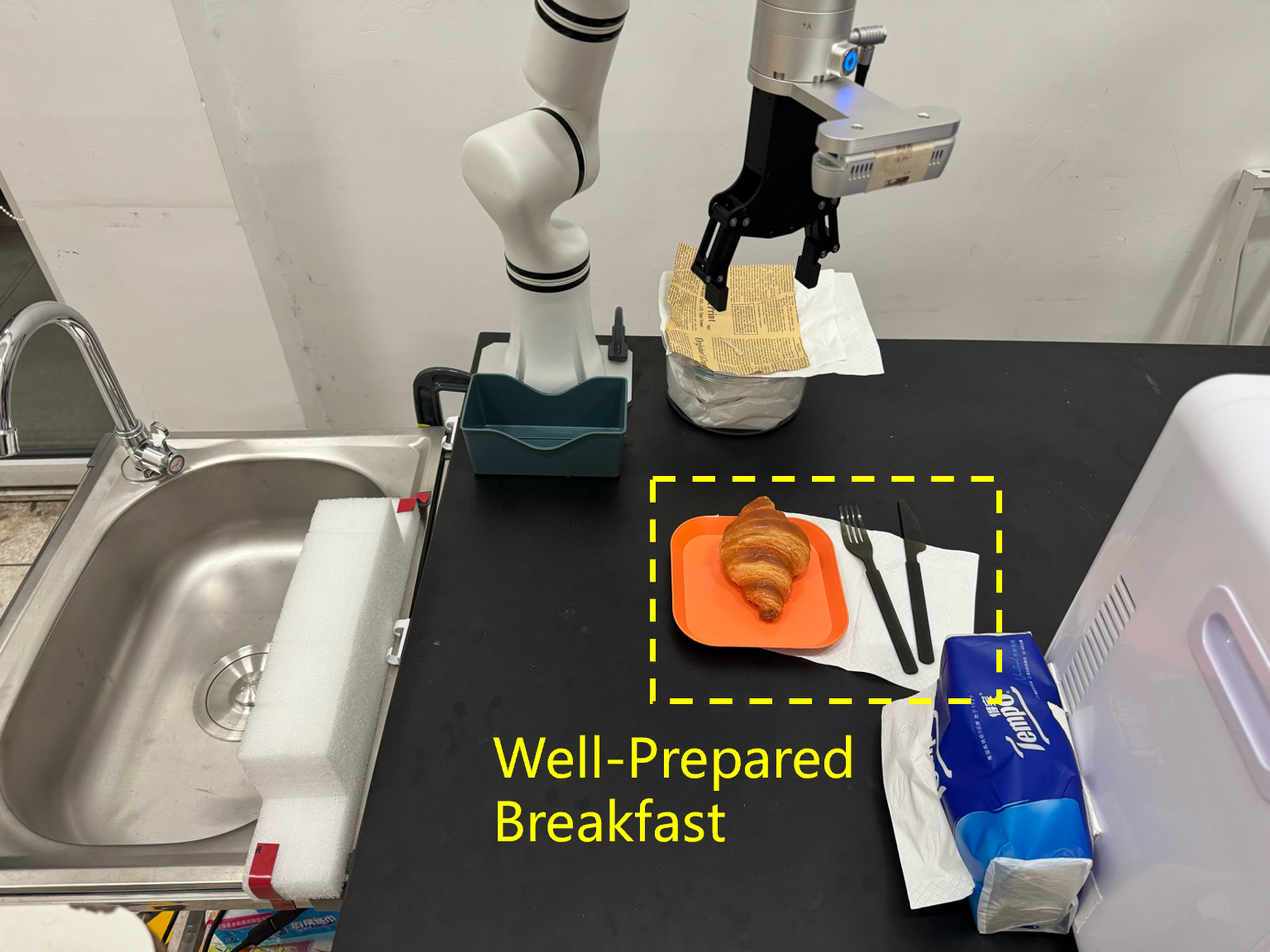}
        \caption{Scene after breakfast task.}
        \label{fig:exp_breakfast_2}
    \end{subfigure}
    \begin{subfigure}[b]{0.27\textwidth}
        \centering
        \includegraphics[width=\textwidth]{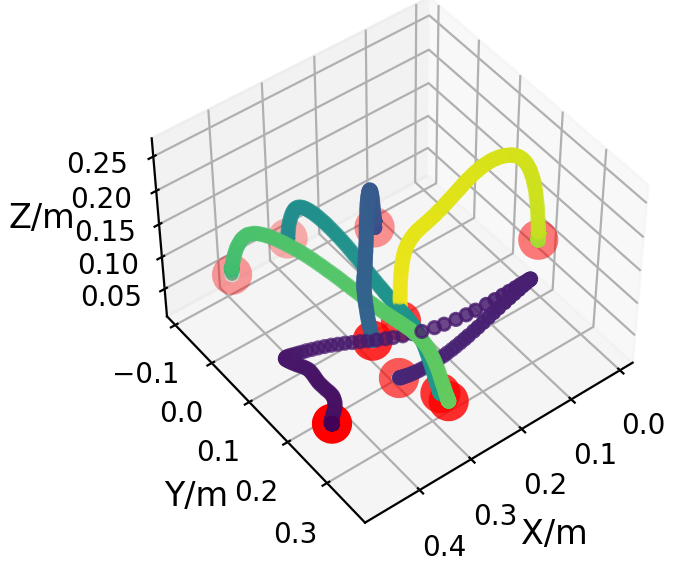}
        \caption{Breakfast task trajectories.}
        \label{fig:exp_breakfast_1}
    \end{subfigure}
    \hspace{0.01\textwidth} 
    \caption{Experiment of breakfast preparation task.}
    \label{fig:exp_breakfast}
\end{figure*}

\begin{figure*}[!htb]
    \centering
    \begin{subfigure}[b]{0.25\textwidth}
        \centering
        \includegraphics[width=\textwidth]{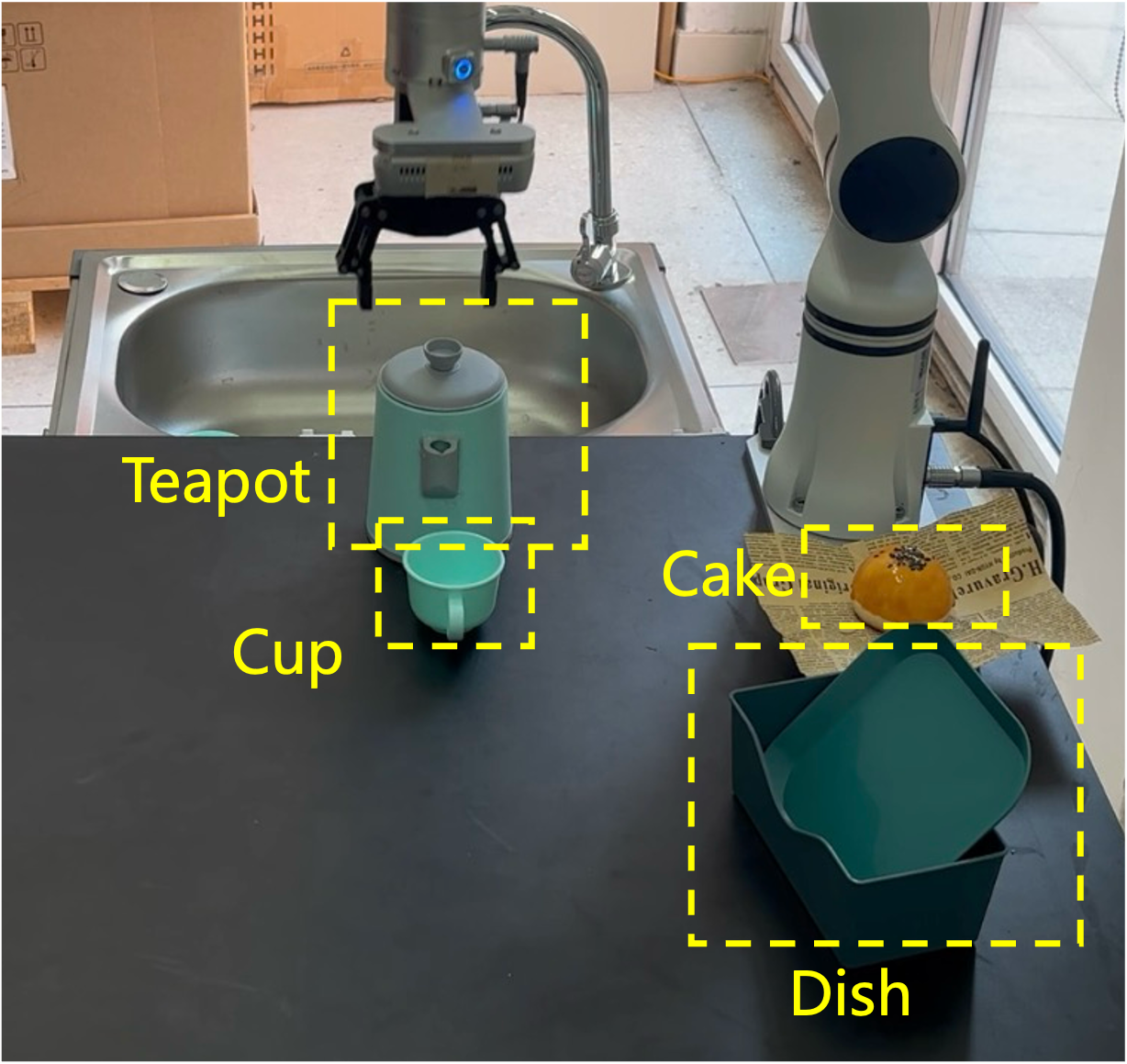}
        \caption{Scene before afternoon tea preparing task.}
        \label{fig:exp_tea_1}
    \end{subfigure}
    \hspace{0.01\textwidth} 
    \begin{subfigure}[b]{0.25\textwidth}
        \centering
        \includegraphics[width=\textwidth]{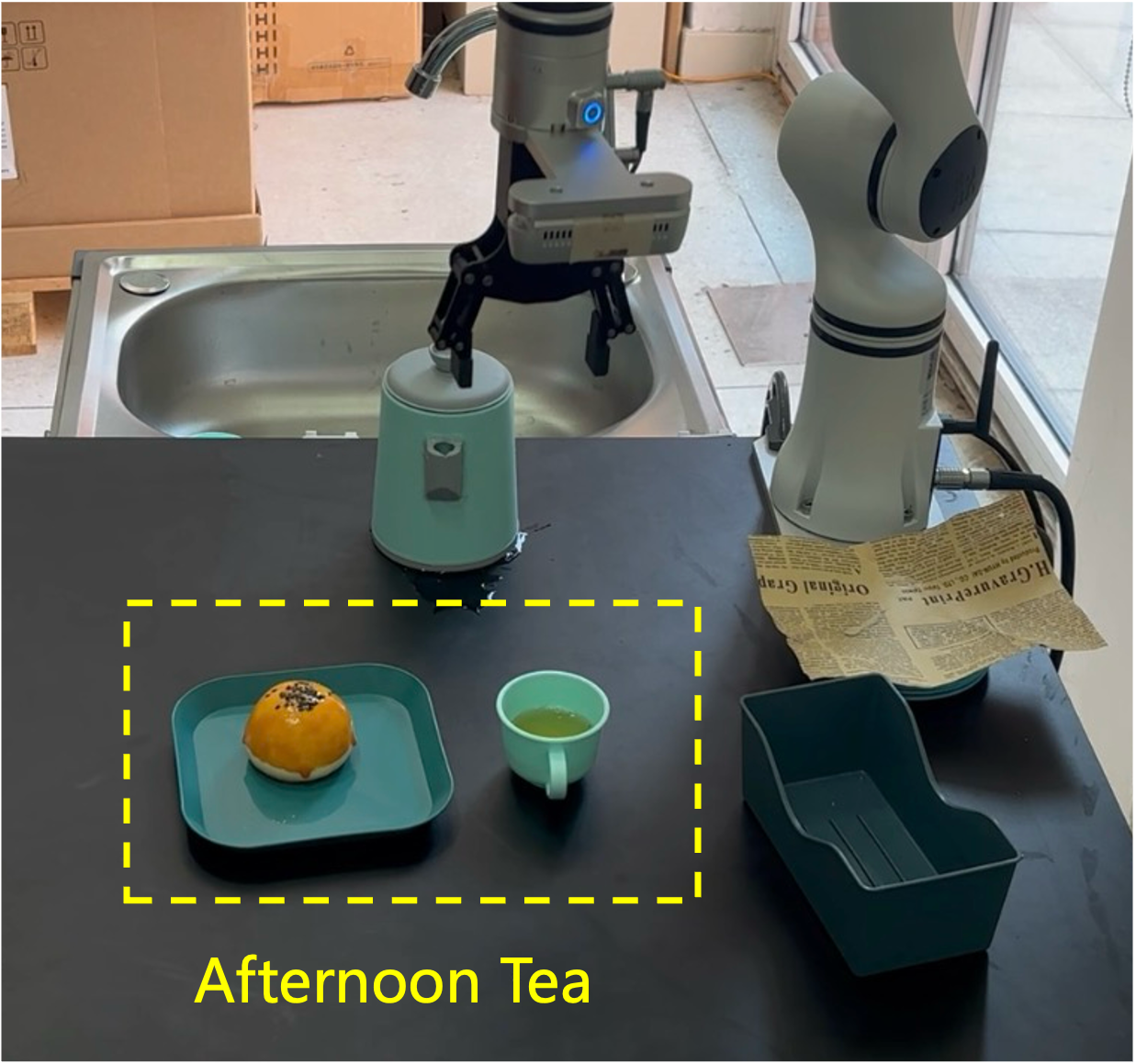}
        \caption{Scene after afternoon tea preparing task.}
        \label{fig:exp_tea_1}
    \end{subfigure}
    \hspace{0.01\textwidth} 
    \begin{subfigure}[b]{0.25\textwidth}
        \centering
        \includegraphics[width=\textwidth]{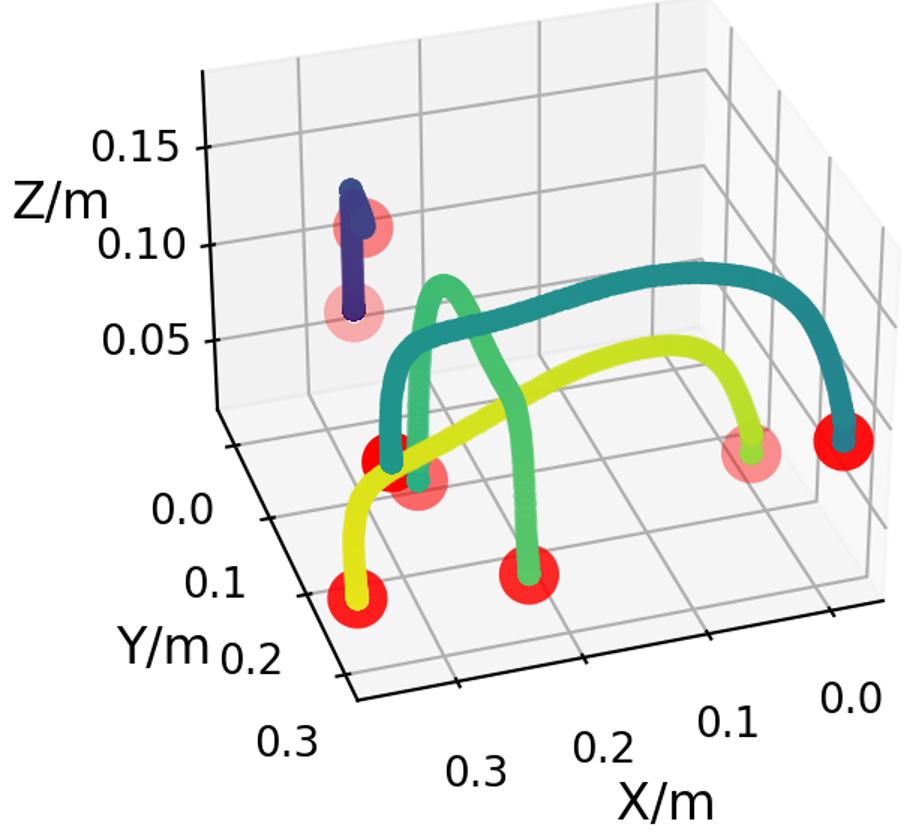}
        \caption{Afternoon tea preparing task trajectories.}
        \label{fig:exp_tea_2}
    \end{subfigure}
    \caption{Experiment of afternoon tea preparation task.}
    \label{fig:exp_tea}
\end{figure*}

Each manipulation skill demonstrated by human operators is captured and recorded by Qualisys motion capture system with eight cameras.
The human teacher performs the manipulation skills with a glove on which four markers are stuck. 
The positions of these markers are calculated to acquire the hand positions and rotations. These acquired demonstrations are then learned by DMPs in Section \ref{subsec:LfD}.
The corresponding motion is encoded using DMPs with both Cartesian and quaternion formulations for position and orientation, respectively. The DMPs were learned from multiple demonstrations of each skill. GMM-GMR method is  applied to encode the variability and generate the distributions, while the means of the trajectories are learned by DMPs.
There are three main skills learned and used in the DMPs bank: \textbf{pick and place object}, \textbf{yank out and spread out napkin}, and \textbf{pour tea}.
The learned DMPs are subsequently optimized with task-specific STL constraints while preserving the dynamic characteristics via forcing-term matching, as illustrated in Section~\ref{subsec:3.3}.

The framework begins by encoding high‐level logic as STL formulas (in Section~\ref{sec:representSTL} and ~\ref{sec:STL2BTs}), which are converted into an LTL representation that underpins the Behavior Tree structure. Each manipulation action is then encapsulated in an atomic BT combining its pre- and post‐conditions with constraints derived from STL. The reactive composition of these nodes enables a hierarchical organization of task execution, facilitating real-time monitoring, automatic re-execution of failed steps, and adaptive satisfaction of spatiotemporal constraints. The BT outputs a subtask signal $\epsilon$ and propagates task and motion‐level constraints $\varphi$ to the motion planner, where DMPs are optimized under STL constraints using the forcing‐term similarity method (Section~\ref{sec:STL}). This ensures that skills learned from multiple demonstrations generalize robustly to complex long‐horizon tasks. Two designed scenarios are shown in the following parts.

\subsection{Scenario 1: Breakfast Preparation}
In this task, the robot is required to prepare a simple breakfast involving multi-step actions such as grasping fork, knife and plate, placing it on the table, and yanking napkin out of napkin box. The manipulation sequence is decomposed into a series of atomic actions with STL-formulated constraints ensuring safety and logical requirements. For instance, the yanking and spread napkin task is restricted by a via-point constraint to spread out the napkin on the table, modeled as ${\Diamond}_{[T_1,T_2]}(\|\textbf{y} - \textbf{y}_\text{via}\|^2 < 0.01)$. Besides, The workspace is limited via ${\square}_{[0,T]}(\left\{ x_{bu} , y_{bu} , z_{bu}  \right\} < \textbf{y} < \left\{ x_{up} , y_{up} , z_{up}  \right\})$ to prevent collisions with surrounding objects.
{As shown in Fig.~\ref{fig:exp_breakfast}, the learned DMPs successfully adapt to the target positions, while satisfying all specified STL constraints. }

\subsection{Scenario 2: Afternoon Tea Preparation}
The second experiment performs an afternoon tea preparation task, consisting of manipulation tasks such as pouring water and placing the dessert when avoiding an obstacle. STL constraints include obstacle avoidance constraints ${\square}_{[0,T]}(d(\textbf{y}, \textbf{y}_{obs} ) > 0.02)$ and velocity limit formulated as ${\square}_{[T_3,T_4]}(\dot{\textbf{y}} < 0.005)\land {\square}_{[T_3,T_4]}(\dot{\textbf{q}} < 0.01)$ to ensure the stability when moving a cup filled with liquid.

{
As shown in Fig.~\ref{fig:exp_tea}, the framework successfully generalize the learned motion primitives to this more delicate context, dynamically adapting to the environment as well as satisfying all logical and physical constraints. The use of STL-constrained DMPs allows motions to maintain the original demonstration dynamics while being robust to variations in spatial layout.}

\section{CONCLUSIONS}

This paper presents a task and motion planning framework that integrates Behavior Trees, Temporal Logic, and Dynamical Movement Primitives to enhance Learning from Demonstration for long-horizon robotic manipulation. STL representations are inputted and transformed into BTs with a remained STL-style constraint. By embedding STL-based spatiotemporal constraints into atomic BT nodes and optimizing DMPs via forcing-term similarity, the framework ensures modularity, reactivity, and generalization across diverse scenarios. The system demonstrates robust performance in both simulation and real-world tasks (e.g., breakfast and afternoon-tea preparation). Future work will focus on perception-guided semantics and large language model-driven temporal logic planning, with the goal of enabling real-time adaptation to novel tasks and enhancing autonomy in constraint-sensitive robotic applications.

\bibliographystyle{bib/IEEEtrans}
\bibliography{main}

\begin{thebibliography}{10}
\providecommand{\url}[1]{#1}
\csname url@samestyle\endcsname
\providecommand{\newblock}{\relax}
\providecommand{\bibinfo}[2]{#2}
\providecommand{\BIBentrySTDinterwordspacing}{\spaceskip=0pt\relax}
\providecommand{\BIBentryALTinterwordstretchfactor}{4}
\providecommand{\BIBentryALTinterwordspacing}{\spaceskip=\fontdimen2\font plus
\BIBentryALTinterwordstretchfactor\fontdimen3\font minus \fontdimen4\font\relax}
\providecommand{\BIBforeignlanguage}[2]{{%
\expandafter\ifx\csname l@#1\endcsname\relax
\typeout{** WARNING: IEEEtran.bst: No hyphenation pattern has been}%
\typeout{** loaded for the language `#1'. Using the pattern for}%
\typeout{** the default language instead.}%
\else
\language=\csname l@#1\endcsname
\fi
#2}}
\providecommand{\BIBdecl}{\relax}
\BIBdecl

\bibitem{garrett2021integrated}
C.~R. Garrett, R.~Chitnis, R.~Holladay, B.~Kim, T.~Silver, L.~P. Kaelbling, and T.~Lozano-P{\'e}rez, ``Integrated task and motion planning,'' \emph{Annual Review of Control, Robotics, and Autonomous Systems}, vol.~4, no.~1, pp. 265--293, 2021.

\bibitem{guo2023recent}
H.~Guo, F.~Wu, Y.~Qin, R.~Li, K.~Li, and K.~Li, ``Recent trends in task and motion planning for robotics: A survey,'' \emph{ACM Computing Surveys}, vol.~55, no. 13s, pp. 1--36, 2023.

\bibitem{zhao2024survey}
Z.~Zhao, S.~Cheng, Y.~Ding, Z.~Zhou, S.~Zhang, D.~Xu, and Y.~Zhao, ``A survey of optimization-based task and motion planning: From classical to learning approaches,'' \emph{IEEE/ASME Transactions on Mechatronics}, pp. 1--27, 2024.

\bibitem{ravichandar2020recent}
H.~Ravichandar, A.~S. Polydoros, S.~Chernova, and A.~Billard, ``Recent advances in robot learning from demonstration,'' \emph{Annual Review of Control, Robotics, and Autonomous Systems}, vol.~3, pp. 297--330, 2020.

\bibitem{lu2021constrained}
Z.~Lu, N.~Wang, and C.~Yang, ``A constrained dmps framework for robot skills learning and generalization from human demonstrations,'' \emph{IEEE/ASME Transactions on Mechatronics}, vol.~26, no.~6, pp. 3265--3275, 2021.

\bibitem{zhou2020incremental}
X.~Zhou, H.~Wu, J.~Rojas, Z.~Xu, S.~Li, X.~Zhou, H.~Wu, J.~Rojas, Z.~Xu, and S.~Li, ``Incremental learning robot task representation and identification,'' \emph{Nonparametric Bayesian Learning for Collaborative Robot Multimodal Introspection}, pp. 29--49, 2020.

\bibitem{davoodi2022rule}
M.~Davoodi, A.~Iqbal, J.~M. Cloud, W.~J. Beksi, and N.~R. Gans, ``Rule-based safe probabilistic movement primitive control via control barrier functions,'' \emph{IEEE Transactions on Automation Science and Engineering}, vol.~20, no.~3, pp. 1500--1514, 2022.

\bibitem{nawaz2024reactive}
F.~Nawaz, S.~Peng, L.~Lindemann, N.~Figueroa, and N.~Matni, ``Reactive temporal logic-based planning and control for interactive robotic tasks,'' in \emph{IEEE/RSJ International Conference on Intelligent Robots and Systems}.\hskip 1em plus 0.5em minus 0.4em\relax IEEE, 2024, pp. 12\,108--12\,115.

\bibitem{colledanchise2018behavior}
M.~Colledanchise and P.~{\"O}gren, \emph{Behavior trees in robotics and AI: An introduction}.\hskip 1em plus 0.5em minus 0.4em\relax CRC Press, 2018.

\bibitem{ogren2022behavior}
P.~{\"O}gren and C.~I. Sprague, ``Behavior trees in robot control systems,'' \emph{Annual Review of Control, Robotics, and Autonomous Systems}, vol.~5, no.~1, pp. 81--107, 2022.

\bibitem{iovino2024comparison}
M.~Iovino, J.~F{\"o}rster, P.~Falco, J.~J. Chung, R.~Siegwart, and C.~Smith, ``Comparison between behavior trees and finite state machines,'' \emph{arXiv preprint arXiv:2405.16137}, 2024.

\bibitem{cloete2024adaptive}
J.~Cloete, W.~Merkt, and I.~Havoutis, ``Adaptive manipulation using behavior trees,'' \emph{arXiv preprint arXiv:2406.14634}, 2024.

\bibitem{leung2023backpropagation}
K.~Leung, N.~Ar{\'e}chiga, and M.~Pavone, ``Backpropagation through signal temporal logic specifications: Infusing logical structure into gradient-based methods,'' \emph{The International Journal of Robotics Research}, vol.~42, no.~6, pp. 356--370, 2023.

\bibitem{chen2023nl2tl}
Y.~Chen, R.~Gandhi, Y.~Zhang, and C.~Fan, ``Nl2tl: Transforming natural languages to temporal logics using large language models,'' \emph{arXiv preprint arXiv:2305.07766}, 2023.

\bibitem{wang2024chatstl}
Y.~Wang, Z.~Huang, S.~Dong, H.~Chu, X.~Yin, and B.~Gao, ``Chatstl: A framework of translation from natural language to signal temporal logic specifications for autonomous vehicle navigation out of blocked scenarios,'' in \emph{International Conference on Computer and Automation Engineering}.\hskip 1em plus 0.5em minus 0.4em\relax IEEE, 2024, pp. 483--487.

\bibitem{yin2024formal}
X.~Yin, B.~Gao, and X.~Yu, ``Formal synthesis of controllers for safety-critical autonomous systems: Developments and challenges,'' \emph{Annual Reviews in Control}, vol.~57, p. 100940, 2024.

\bibitem{iovino2023programming}
M.~Iovino, J.~F{\"o}rster, P.~Falco, J.~J. Chung, R.~Siegwart, and C.~Smith, ``On the programming effort required to generate behavior trees and finite state machines for robotic applications,'' in \emph{IEEE International Conference on Robotics and Automation}.\hskip 1em plus 0.5em minus 0.4em\relax IEEE, 2023, pp. 5807--5813.

\bibitem{wang2024novel}
W.~Wang, C.~Zeng, H.~Zhan, and C.~Yang, ``A novel robust imitation learning framework for complex skills with limited demonstrations,'' \emph{IEEE Transactions on Automation Science and Engineering}, 2024.

\bibitem{dominguez2025beyond}
D.~C. Dom{\'\i}nguez, E.~Schaffernicht, and T.~Stoyanov, ``Beyond predefined actions: Integrating behavior trees and dynamic movement primitives for robot learning from demonstration,'' \emph{arXiv preprint arXiv:2505.08625}, 2025.

\bibitem{plaku2015motion}
E.~Plaku and S.~Karaman, ``Motion planning with temporal-logic specifications: Progress and challenges,'' \emph{AI Communications}, vol.~29, no.~1, pp. 151--162, 2015.

\bibitem{aeronautiques1998pddl}
C.~Aeronautiques, A.~Howe, C.~Knoblock, I.~D. McDermott, A.~Ram, M.~Veloso, D.~Weld, D.~W. Sri, A.~Barrett, D.~Christianson \emph{et~al.}, ``Pddl| the planning domain definition language,'' \emph{Technical Report, Tech. Rep.}, 1998.

\bibitem{van2024vernacopter}
T.~V. Laar, Z.~Zhang, S.~Qi, S.~Haesaert, and Z.~Sun, ``Vernacopter: Disambiguated natural-language-driven robot via formal specifications,'' \emph{arXiv preprint arXiv:2409.09536}, 2024.

\bibitem{lindemann2018control}
L.~Lindemann and D.~V. Dimarogonas, ``Control barrier functions for signal temporal logic tasks,'' \emph{IEEE Control Systems Letters}, vol.~3, no.~1, pp. 96--101, 2018.

\bibitem{saveriano2023dynamic}
M.~Saveriano, F.~J. Abu-Dakka, A.~Kramberger, and L.~Peternel, ``Dynamic movement primitives in robotics: A tutorial survey,'' \emph{The International Journal of Robotics Research}, vol.~42, no.~13, pp. 1133--1184, 2023.

\bibitem{liu2025novel}
Z.~Liu and Y.~Fang, ``A novel dmps framework for robot skill generalizing with obstacle avoidance: Taking volume and orientation into consideration,'' \emph{IEEE/ASME Transactions on Mechatronics}, 2025.

\bibitem{meli2023logic}
D.~Meli, H.~Nakawala, and P.~Fiorini, ``Logic programming for deliberative robotic task planning,'' \emph{Artificial Intelligence Review}, vol.~56, no.~9, pp. 9011--9049, 2023.

\bibitem{yu2024continuous}
P.~Yu, X.~Tan, and D.~V. Dimarogonas, ``Continuous-time control synthesis under nested signal temporal logic specifications,'' \emph{IEEE Transactions on Robotics}, vol.~40, pp. 2272--2286, 2024.

\bibitem{cardona2023flexible}
G.~A. Cardona, K.~Leahy, M.~Mann, and C.-I. Vasile, ``A flexible and efficient temporal logic tool for python: Pytelo,'' \emph{arXiv preprint arXiv:2310.08714}, 2023.

\bibitem{mao2024nl2stl}
Y.~Mao, T.~Zhang, X.~Cao, Z.~Chen, X.~Liang, B.~Xu, and H.~Fang, ``Nl2stl: Transformation from logic natural language to signal temporal logics using llama2,'' in \emph{IEEE International Conference on Cybernetics and Intelligent Systems and IEEE International Conference on Robotics, Automation and Mechatronics}.\hskip 1em plus 0.5em minus 0.4em\relax IEEE, 2024, pp. 469--474.

\bibitem{choe2025seeing}
D.~B. Choe, S.~V. Sangeetha, S.~Emanuel, C.-Y. Chiu, S.~Coogan, and S.~Kousik, ``Seeing, saying, solving: An llm-to-tl framework for cooperative robots,'' \emph{arXiv preprint arXiv:2505.13376}, 2025.

\bibitem{fang2025enhancing}
Y.~Fang, Z.~Jin, J.~An, H.~Chen, X.~Chen, and N.~Zhan, ``Enhancing transformation from natural language to signal temporal logic using llms with diverse external knowledge,'' \emph{arXiv preprint arXiv:2505.20658}, 2025.

\bibitem{neupane2023designing}
A.~Neupane, E.~G. Mercer, and M.~A. Goodrich, ``Designing behavior trees from goal-oriented ltlf formulas,'' \emph{arXiv preprint arXiv:2307.06399}, 2023.

\bibitem{pek2023spatial}
C.~Pek, G.~F. Schuppe, F.~Esposito, J.~Tumova, and D.~Kragic, ``Spatial: monitoring and planning of robotic tasks using spatio-temporal logic specifications,'' \emph{Autonomous Robots}, vol.~47, no.~8, pp. 1439--1462, 2023.

\end{thebibliography}

\appendix

\subsection{Construction for the Weight of Optimization Function in Generalizing Multi-demon-learned DMPs}\label{sec:OFC}

{
As described by equation~\eqref{eq:dmp_objective_2} in Section~\ref{subsec:3.3}, this formulation offers distinct advantages over direct trajectory optimization approaches. 
The use of means and variances obtained from learning from multiple demonstrations (Section~\ref{subsec:LfD}) enables adaptive weighting of the forcing term optimization based on trajectory variance analysis, allowing the robot to maintain effective dynamics in key trajectory segments—specifically, those with lower variance.
}

{For the coherence and consistency of the text, we take \textit{3-Dimension case}, meaning the 3D trajectory of the DMPs to continue the following expression: $\mathbf{F^{\epsilon}} = \{ \mathbf{F}_x^{\epsilon}, \mathbf{F}_y^{\epsilon}, \mathbf{F}_z^{\epsilon} \}$.}

\begin{remark}
    {
    For the pose case, the forcing term is represented as $\mathbf{F}^{\epsilon} = \{ \mathbf{F}_x^{\epsilon}, \mathbf{F}_y^{\epsilon}, \mathbf{F}_z^{\epsilon}, \mathbf{F}_u^{\epsilon}, \mathbf{F}_a^{\epsilon}, \mathbf{F}_b^{\epsilon}, \mathbf{F}_c^{\epsilon} \}$, where $u$, $a$, $b$, and $c$ are the scalar and vector components of the quaternion defined as $\mathbf{q} = u + a\hat{\imath} + b\hat{\jmath} + c\hat{k}$. 
    In this section, we focus on the 3-dimensional case of $\mathbf{F}^{\epsilon}$ for the sake of clarity and concise formula representation.
    }
\end{remark}

Gained from the GMM-GMR, the variances of the demo trajectories can be written as:
\begin{equation}
    \begin{aligned}
        \Sigma &= [\Sigma_1, \Sigma_2, ..., \Sigma_t, ...,\Sigma_T], 
        \\
        \Sigma_t &= 
        \begin{bmatrix}
        \sigma^t_{xx} & \sigma^t_{xy} & \sigma^t_{xz} \\
        \sigma^t_{yx} & \sigma^t_{yy} & \sigma^t_{yz} \\
        \sigma^t_{zx} & \sigma^t_{zy} & \sigma^t_{zz} \\
        \end{bmatrix}, \forall t=1,2,...,T,
    \end{aligned}
\end{equation}
where $\sigma$ represents the variance of the demonstrations, $t$ is one of the time step in the time range $[1,...,T]$. These variances can be used to facilitate the generalization of the motions, by giving a weight of the forcing function both in time and space dimensions. The specific formulas are introduced as follows:

For the variance weight in time-variance level:
\begin{equation}
    \begin{aligned}
        \mathcal{W}_{\delta\delta} &= [\omega_{\delta\delta}^1, \omega_{\delta\delta}^2, ..., \omega_{\delta\delta}^i ,..., \omega_{\delta\delta}^T], \quad 
   \forall \delta \in \{x, y, z\}, \\
      \omega_{\delta\delta}^i &= 1/e^{\frac{\sigma^i_{\delta\delta}}{(\sum_{i=0}^T{\sigma_{\delta\delta}^i})/T}}, \quad 
      \forall i=1,2,...,T.
    \end{aligned}
\end{equation}

For the variance weight in space-variance level:
\begin{equation}
    \begin{aligned}
      \mathcal{W}_t &= [\omega_{t}^{x}, \omega_{t}^{y}, \omega_{t}^{z}],  \quad
     \omega_{t}^{\delta}  = 1/e^{\frac{\sigma_{t}^\delta}{(\sigma_{t}^x+ \sigma_{t}^y+ \sigma_{t}^z)/3}},   
      \forall  \delta \in \{x, y, z\}.        
    \end{aligned}
\end{equation}

It is remarkable that,
DMPs require all demonstrations to have the same start and goal positions (or be scaled to be consistent), so as to leverage the convergence properties of DMPs and learn the dynamic characteristics from multiple demonstrations. The same start and goal positions bring the 0 variance value at start and goal points, which will cause further an infinite value when calculating the weight. Hence, the \textit{exponential function structures} are introduced in the above equations to avoid this problem.  
Then, the variance weight $\mathcal{W}$ in equation \eqref{eq:dmp_objective} can be represented by:
\begin{equation}\label{eq:matcalW}
    \mathcal{W} = \mathcal{W}_t \cdot 
    \begin{bmatrix}
    \mathcal{W}_{xx}\\ \mathcal{W}_{yy}\\ \mathcal{W}_{zz}
    \end{bmatrix}.
\end{equation}

With the weight matrix in \eqref{eq:matcalW}, the DMPs optimization function $J_{\text{DMP}}$ can be enhanced as equation \eqref{eq:dmp_objective}.

\end{document}